\documentclass[runningheads]{llncs}

\usepackage{eccv}
\usepackage{eccvabbrv}

\usepackage{hyperref}

\usepackage{amsmath}
\usepackage{amssymb}
\usepackage{booktabs}
\usepackage{graphicx}
\usepackage{multirow}
\usepackage{siunitx}
\usepackage{verbatim}
\usepackage{algpseudocode}
\usepackage{listings}
\usepackage[font=small]{caption}

\usepackage[capitalize]{cleveref}
\crefname{section}{Sec.}{Secs.}
\Crefname{section}{Section}{Sections}
\Crefname{table}{Table}{Tables}
\crefname{table}{Tab.}{Tabs.}

\usepackage{etoolbox} 

\makeatletter
\patchcmd{\maketitle}
  {\def\thanks}
  {\let\repthanks\repthanksunskip\let\repthx\repthxunskip\let\reprepthx\reprepthxunskip\def\thanks}
  {}{}
\patchcmd{\@maketitle}
  {\def\thanks}
  {\let\repthanks\@gobble\let\repthx\@gobble\let\reprepthx\@gobble\def\thanks}
  {}{}

\newcommand\repthanksunskip[1]{\unskip{}}

\newcommand\repthxunskip[1]{\unskip{}}

\newcommand\reprepthxunskip[1]{\unskip{}}

\newcommand{\repthanks}[1]{\textsuperscript{\dag}}

\newcommand{\repthx}[1]{\textsuperscript{\ddag}}

\newcommand{\reprepthx}[1]{\textsuperscript{\S}}

\makeatother

\authorrunning{Qin et al.}

\begin{document}

\lstset{
    language=Python,
    basicstyle=\small\ttfamily,
    commentstyle=\color{green!50!black},
    keywordstyle=\color{blue},
    stringstyle=\color{red},
    showstringspaces=false,
    frame=single
}
\title{MobileNetV4: Universal Models for the Mobile Ecosystem}

\author{Danfeng Qin\repthanks{Key Contributor\protect\label{key-contributor}}\repthx{Project Lead.\protect\label{project-lead}} \and
Chas Leichner \repthanks{key-contributor}\repthx{project-lead} \and
Manolis Delakis \repthanks{key-contributor} \and
Marco Fornoni \and
Shixin Luo \and Fan Yang \and Weijun Wang \and 
Colby Banbury  \and Chengxi Ye \and Berkin Akin
\and Vaibhav Aggarwal \and Tenghui Zhu \and Daniele Moro
\and Andrew Howard\repthanks{key-contributor}\reprepthx{Senior Lead.\protect\label{senior-lead}}
} 

\institute{Google}

\maketitle

\footnotetext[0]{\dag Equal primary contribution. \ddag Project Lead. \S Senior Lead.}

\begin{abstract}
We present the latest generation of MobileNets: MobileNetV4 (MNv4). They feature universally-efficient architecture designs for mobile devices.
We introduce the Universal Inverted Bottleneck (UIB) search block, a unified and flexible structure that merges Inverted Bottleneck (IB), ConvNext, Feed Forward Network (FFN), and a novel Extra Depthwise (ExtraDW) variant. 
Alongside UIB, we present Mobile MQA, an attention block for mobile accelerators, delivering a significant 39\% speedup.
An optimized neural architecture search (NAS) recipe is also introduced which improves MNv4 search effectiveness. 
The integration of UIB, Mobile MQA and the refined NAS recipe results in a new suite of MNv4 models that are mostly Pareto optimal across mobile CPUs, DSPs, GPUs, as well as accelerators like Apple Neural Engine and Google Pixel EdgeTPU.
This performance uniformity is not found in any other models tested.
We introduce performance modeling and analysis techniques to explain how this performance is achieved.
Finally, to further boost accuracy, we introduce a novel distillation technique. 
Enhanced by this technique, our MNv4-Hybrid-Large model delivers 87\% ImageNet-1K accuracy, with a Pixel 8 EdgeTPU runtime of 3.8ms.
\end{abstract}

\section{Introduction}
\label{sec:intro}
Efficient on-device neural networks not only enable fast, real-time and interactive experiences, but also avoid streaming private data through the public internet. 
However, the computational constraints of mobile devices pose the significant challenge of balancing accuracy and efficiency.
To this end, we introduce UIB and Mobile MQA, two innovative building blocks integrated via a refined NAS recipe to create a series of universally mostly-Pareto-optimal mobile models.\footnote{All MobileNetV4 models are available at {\scriptsize \url{https://github.com/tensorflow/models/blob/master/official/vision/modeling/backbones/mobilenet.py}}}
Additionally, we present a distillation technique that further improves efficiency.

Our Universal Inverted Bottleneck (UIB) block improves the Inverted Bottleneck block~\cite{mobilenet_v2} by incorporating two optional depthwise convolutions~\cite{mobilenet}. 
Despite its simplicity, UIB unifies prominent micro-architectures - Inverted Bottleneck (IB), ConvNext~\cite{convnext}, and FFN~\cite{dosovitskiy2020image} - and introduces the Extra Depthwise (ExtraDW) IB block. 
UIB offers flexibility in spatial and channel mixing, the option to extend the receptive field, and enhanced computational efficiency.

Our optimized Mobile MQA block achieves over a 39\% inference speedup on mobile accelerators with respect to Multi-Head Attention~\cite{vaswani2017attention}.

\begin{figure}[t]
    \centering 
    \includegraphics[width=\columnwidth]{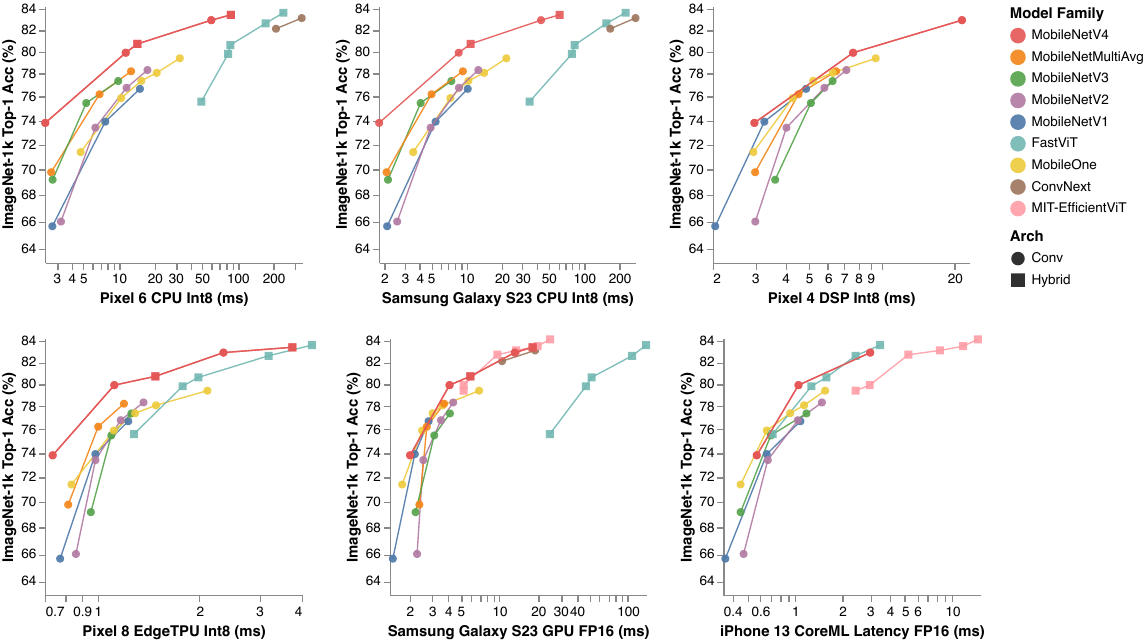}
\caption{\small \textbf{MNv4 Models are Universally Mostly Pareto Optimal:} 
MNv4 performs strongly compared to leading efficient models across diverse hardware.
All models were trained on ImageNet-1k solely.
MobileNetV1-V3 were retrained with updated recipes.
Most models were optimized for one device, but MNv4 is Pareto optimal across most devices.
Hybrid models and ConvNext are DSP-incompatible.
Due to PyTorch-to-TFLite export tool limitations, EfficientViTs~\cite{mit-efficientvit}~\cite{msr-efficientvit} are not benchmarked on CPUs and EdgeTPU.
MNv4-Hybrid models were excluded from CoreML evaluation due to the lack of PyTorch implementation of Mobile MQA.
}
\label{fig:multi_hardware_pareto}
\end{figure}

Our two-phase NAS approach, separating coarse and fine-grained searches, significantly boosts search efficiency and facilitates the creation of models that are significantly larger than previous state-of-the-art models~\cite{tan2019efficientnet}. 
Additionally, incorporating an offline distillation dataset reduces noise in NAS reward measurements, resulting in improved model quality.

By integrating UIB, MQA, and an improved NAS recipe, we present the MNv4 suite of models which achieve mostly Pareto optimal performance across diverse hardware platforms, including CPUs, DSPs, GPUs, and accelerators. 
Our models range from the extremely compact MNv4-Conv-S to the MNv4-Hybrid-L high-end variant that establishes a new reference for mobile model accuracy.
MNv4-Conv-S achieves 73.8\% top-1 ImageNet-1K accuracy with 3.8M parameters, 0.2G MACs and 2.4 ms of Pixel 6 CPU latency.  MNv4-Hybrid-L gets 83.4\% top-1 within 3.8 ms on Pixel 8 EdgeTPU.
Our novel distillation recipe mixes datasets with different augmentations and adds balanced in-class data, enhancing generalization and increasing accuracy. 
With these techniques, MNv4-Hybrid-L achieves a 87\% top-1 accuracy on ImageNet-1K: 0.5\% less than its teacher, despite having 39x less MACs.

\section{Related Work}
\label{sec:related}
Optimizing models for both accuracy and efficiency is a well studied problem.

\textbf{Mobile Convolutional Networks:} Key work includes MobileNet\-V1~\cite{howard2017mobilenets} with depthwise-separable convolutions for better efficiency, MobileNet\-V2~\cite{mobilenet_v2} introducing linear bottlenecks and inverted residuals, GhostNet~\cite{han2020ghostnet} increasing the relative frequency of depthwise convolutions, MnasNet~\cite{mnas} integrating lightweight attention in bottlenecks, and MobileOne~\cite{vasu2023mobileone} adding and re-parameterizing linear branches in inverted bottlenecks at inference time.

\textbf{Efficient Hybrid Networks:} This research combines convolutions and attention. 
MobileViT~\cite{mehta2021mobilevit} merges CNN strengths with ViT~\cite{dosovitskiy2020image} through global attention blocks.
GhostNetV2~\cite{tang2022ghostnetv2} uses FC layers to capture long-range dependencies.
MobileFormer~\cite{mobileformer} parallelizes a MobileNet and a Transformer with a two-way bridge in between for feature fusing.
FastViT~\cite{vasu2023fastvit} adds attention to the last stage with large convolutional kernels instead of early stage self-attention.

\textbf{Efficient Attention:} Research has focused on enhancing MHSA~\cite{vaswani2017attention} efficiency. EfficientViT~\cite{mit-efficientvit} and MobileViTv2~\cite{mehta2022mobilevitv2} introduce self-attention approximations for linear complexity with minor accuracy impacts. EfficientFormer\-V2~\cite{efficientformerv2} downsamples Q, K, and V for efficiency, while CMT~\cite{guo2022cmt} and NextViT~\cite{li2022nextvit} downsample only K and V.

\textbf{Hardware-aware Neural Architecture Search (NAS): }
Another common technique is to automate the model design process using hardware-aware Neural Architecture Search (NAS).
NetAdapt~\cite{netadapt} uses empirical latency tables to optimize the accuracy of a model under a target latency constraint. 
MnasNet~\cite{mnas} also uses latency tables, but applies reinforcement learning to do hardware-aware NAS. 
FBNet~\cite{fbnet} accelerates multi-task hardware-aware search via differentiable NAS. 
MobileNetV3~\cite{mobilenet_v3} is tuned to mobile phone CPUs through a combination of hardware-aware NAS, the NetAdapt algorithm, and architecture advances. 
MobileNet MultiHardware~\cite{mobilenet_multi} optimizes a single model for multiple hardware targets. 
Once-for-all~\cite{cai2019once} separates training and search for efficiency.

\section{Hardware-Independent Pareto Efficiency}
\begin{figure}
    \centering 
    \includegraphics[width=\columnwidth]{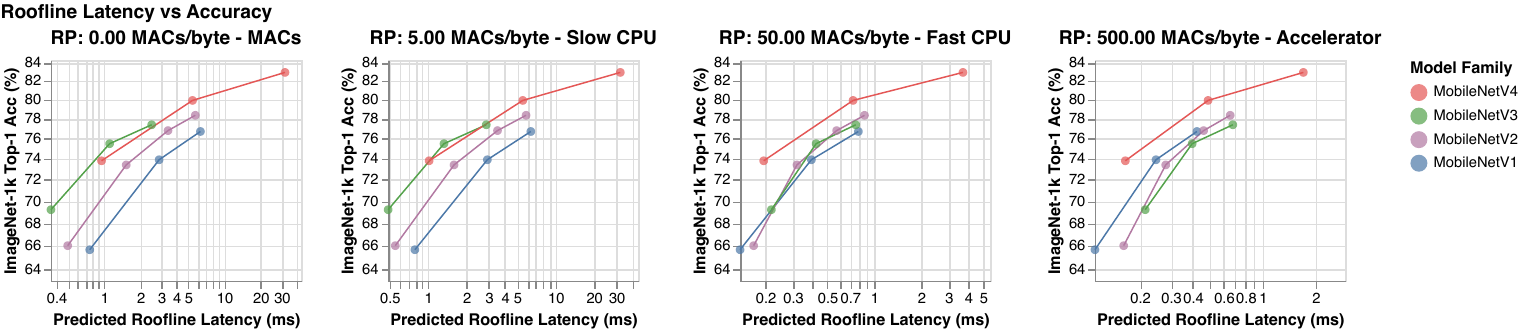} 
\caption{\small \textbf{Ridge Points and Latency/Accuracy Trade-Offs}:
In the roofline performance model, the ridge point summarizes the relationship between memory bandwidth and MACs.
If memory bandwidth is constant, high-compute hardware (accelerators) have a higher ridge point than low-compute hardware (CPUs).
MobileNetV4 is mostly Pareto-optimal from a ridge point of 0 to 500 MACs/byte. 
These analytically-derived (\cref{eq:roofline_model}) charts reflect the real hardware measurements in \cref{fig:multi_hardware_pareto}. \cref{appendix:universality} contains further analysis of this relationship.
}
\label{fig:hardware-independent-pareto}
\end{figure}

\textbf{The Roofline Model}: For a model to be universally efficient, it must perform well on hardware targets with vastly different bottlenecks that limit the model's performance.
These bottlenecks are largely determined by the hardware's peak computational throughput and its peak memory bandwidth.

To this end, we use the Roofline Model~\cite{williams2009roofline} which estimates the performance of a given workload and predicts whether it is memory-bottlenecked or compute-bottlenecked.
In short, it abstracts away specific hardware details and only considers a workload's operational intensity ($\text{LayerMACs}_i/(\text{WeightBytes}_i + \text{ActivationBytes}_i)$) \vs the theoretical limits of the hardware's processor and memory system.
Memory and compute operations happen roughly in parallel, so the slower of the two approximately determines the latency bottleneck.
To apply the Roofline Model to neural networks with layers indexed by $i$, we can calculate the model inference latency, $\text{ModelTime}$, as follows:

\begin{equation}
\small
\begin{aligned}
\text{ModelTime} &= \sum_{i}\max(\text{MACTime}_i, \text{MemTime}_i)\\
  \text{MACTime}_i = \frac{\text{LayerMACs}_i}{\text{PeakMACs}},& \quad \text{MemTime}_i = \frac{\text{WeightBytes}_i + \text{ActivationBytes}_i}{\text{PeakMemBW}}
\end{aligned}
\label{eq:roofline_model}
\end{equation}

\begin{figure}[t!]
    \centering 
    \includegraphics[width=\columnwidth]{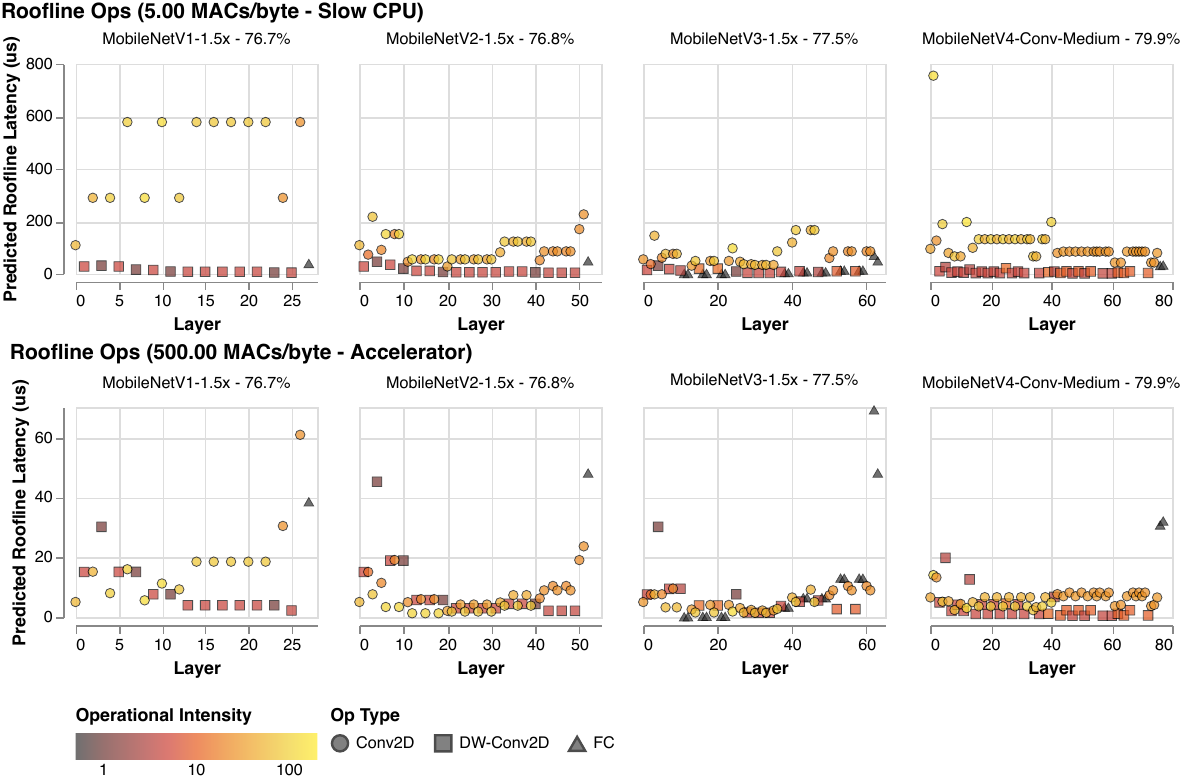} 
\setlength{\abovecaptionskip}{-5pt}
\caption{\textbf{Op Cost \vs Ridge Point}:
Each sub-chart displays the roofline latency (\cref{eq:roofline_model}) of a network's ops.
Networks start on the left.
Large Conv2Ds are expensive on low ridge point (RP) hardware (\textit{top row}), but add cheap model capacity on high-RP hardware (\textit{bottom row}).
FC layers and DW-Conv2Ds are cheap at low RPs and expensive at high RPs.
MobileNetV4 balances MAC-intensive Conv2D layers and memory-intensive FC layers where they contribute most to the network---the beginning and end, respectively. Full sweeps and data for all MobileNetV4-Conv models are in \cref{appendix:universality}.
}
\label{fig:op-cost-vs-ridge-point}
\end{figure}

In the roofline model, hardware behavior is summarized by the \textit{Ridge Point} (RP)---the ratio of a hardware's $\text{PeakMACs}$ to $\text{PeakMemBW}$ \ie the minimum operational intensity required to achieve maximum performance.
\footnote{
The common practice of using a model's total MACs to proxy latency is the same as targeting a roofline model with a Ridge Point $\text{(RP)}=0$.
This is equivalent to infinite bytes per MAC so, $\forall i, \text{MemTime}_i = 0$ and $\text{ModelTime} = \sum_{i}\text{MACTime}_i$.}
In order to optimize for hardware with a wide range of bottlenecks, as seen in \cref{fig:hardware-independent-pareto} and \cref{fig:op-cost-vs-ridge-point}, we analyze our algorithms' latency while sweeping the RP from its lowest expected value (0 MAC/byte) to its highest expected value (500 MACs/byte)---see \cref{appendix:universality} for more details.
Roofline Models only depend on the ratio of data transfer to compute, so all hardware with the same RP will rank workloads the same by latency.\footnote{The Roofline Model assumes that software implementation has no impact on workload performance.
This means techniques with complex memory access (e.g. pruning) perform much better on a Roofline Model than on a real device.}
This means that swept-RP roofline analysis (see next paragraph) applies to future hardware and software if the RP of the new targets is contained in the swept range.\footnote{Andrew Lavin independently proposed a similar framework in the context of analyzing strategies for performance modeling and kernel execution~\cite{lavin2024efficiencyconvolutionalneuralnetworks}.}
 
\textbf{Ridge Point Sweep Analysis}: As seen in \cref{fig:hardware-independent-pareto} and \cref{fig:op-cost-vs-ridge-point}, the roofline model sheds light on how MobileNetV4 models achieve hardware-independent mostly-Pareto-optimal performance against other convolutional MobileNets.
On low-RP hardware (e.g. CPUs), models are more likely to be compute-bound than memory-bound. 
So, to improve latency, you minimize the total number of MACs even at the cost of increased memory complexity (MobileNetV3Large-1.5x).
Data movement is the bottleneck on high-RP hardware, so MACs do not meaningfully slow down the model but can increase model capacity (MobileNetV1-1.5x).
So models optimized for low-RPs run slowly at high-RPs because memory-intensive and low-MAC fully-connected (FC) layers are bottlenecked on memory bandwidth and can't take advantage of the high available $\text{PeakMACs}$.

\textbf{MobileNetV4 Design}: MobileNetV4 balances investing MACs and memory bandwidth where they will provide the maximum return for the cost, paying particular attention to the start and end of the network.
At the beginning of the network, MobileNetV4 uses large and expensive initial layers to substantially improve the models' capacity and downstream accuracy. These initial layers are dominated by a high number of MACs, so they are only expensive on low-RP hardware. 
At the end of the network, all MobileNetV4 variants use the same size final FC layers to maximize accuracy, even though this causes smaller MNV4 variants to suffer higher FC latency on high-RP hardware.  
Since large initial Conv layers are expensive on low-RP hardware but not high-RP hardware while the final FC layers are expensive on high-RP hardware but not low-RP hardware, MobileNetV4 models will never see both slowdowns at the same time. In other words, MNv4 models are able to use expensive layers that disproportionately improve accuracy but do not suffer the simultaneous combined costs of the layers, resulting in mostly Pareto-optimal performance at all ridge points.

\section{Universal Inverted Bottlenecks}
\begin{figure}[t] 
    \centering 
    \includegraphics[width=1.0\columnwidth]{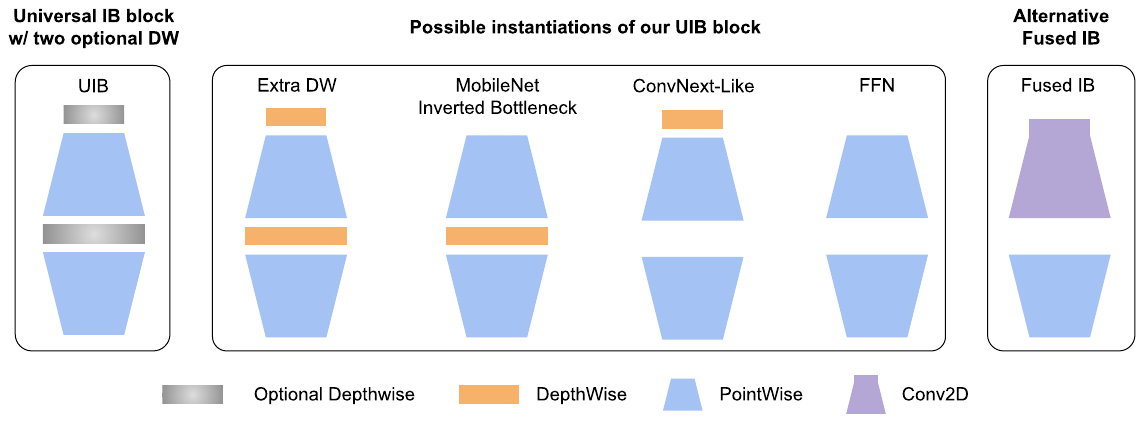} 
    \caption{Universal Inverted Bottleneck (UIB) blocks.} 
    \label{fig:uib_options}
\end{figure}

With an established foundation of roofline modeling and operational intensity, we proceed to discuss our architectural blocks. First is the Universal Inverted Bottleneck (UIB) Block, a building block for efficient network design that can adapt to a variety of optimization targets while remaining simple enough to use with Neural Architecture Search (NAS).
\Cref{fig:uib_options} shows the UIB block structure.

UIB extends the MobileNet Inverted Bottleneck (IB) block (introduced in MobileNetV2~\cite{mobilenet_v2}), which has become the standard building block for efficient networks~\cite{mobilenet_v3,tan2019efficientnet, convnext,dosovitskiy2020image}.
We introduce an optional DW before the expansion layer and also make the DW between the expansion and projection layer optional. The NAS procedure selects which DW ops to include, resulting in novel architectures. 
Despite the simplicity of this modification, our new building block unifies important existing blocks: the original IB block, ConvNext block, and the FFN block in ViT.
Additionally, UIB introduces a novel variant: the Extra DepthWise IB (ExtraDW) block.
The NAS SuperNet size is manageable because the pointwise expansion and projection components of each block are shared between instantions and the depthwise ops are searchable options. In a SuperNet-based NAS algorithm, this approach shares $>$95\% of the parameters between instantiations so NAS remains efficient.
We further use FusedIBs to improve the efficiency:
A $k{\times}k$ FusedIB is a $k{\times}k$ Conv2D into a $1{\times}1$ Conv2D \cite{on_device_neural_archs}. FusedIBs are used in all MNV4 model stems (\cref{appendix:model_detail}, Tabs. \ref{arch_0p5x} - \ref{arch_8x_max}).

\noindent \textbf{UIB Instantiations:} The two optional depthwise convolutions in the UIB block have four possible instantiations (\cref{fig:uib_options}), resulting in different tradeoffs.

\textbf{MobileNet Inverted Bottleneck (IB)} performs spatial mixing on the expanded features' activations for greater model capacity at increased cost.

\textbf{ConvNext-Like} allows for a cheaper spatial mixing with larger kernel size by performing the spatial mixing before the expansion.

\textbf{ExtraDW} inexpensively increases the network depth and receptive field, combining the benefits of ConvNext-Like and IB. \footnote{ExtraDW could be seen as a MobileNetV1-style factorization of two standard convolutional blocks.}

\textbf{FFN} is a stack of two 1x1 pointwise convolutions (PW) with activation and normalization layers in between. PW is very accelerator-friendly but works best with other blocks.

At each network stage, UIB provides flexibility to: (1) Strike an ad-hoc spatial and channel mixing tradeoff.
(2) Enlarge the receptive field as needed. (3) Maximize the computational utilization. 
\cref{tab:block_comparison} shows impact on accuracy and latency across three searches.

\begin{table}[t]
    \centering
    \scriptsize
    \caption{Comparison between searches using Inverted Bottleneck blocks, ConvNext-Like blocks, and full UIB blocks.}
    \begin{tabular}{l c c c c }
    \toprule
    Block        & Top-1           & GMACs        & MParams       & P8 EdgeTPU        \\
    \midrule
    \textbf{UIB} & \textbf{83.3\%} & \textbf{6.2} & \textbf{33.0} & \textbf{2.68 ms}  \\
    CN           & 83.2\%          & 6.9          & 35.1          & 2.69 ms           \\
    IB           & 82.3\%          & 6.1          & 32.4          & 2.61 ms           \\
    \bottomrule
    \end{tabular}
\label{tab:block_comparison}
\end{table}

\section{Mobile MQA}
In this section we present Mobile MQA, a novel accelerator-optimized attention block which speeds up attention by $>$39\%.

\noindent{\textbf{Importance of Operational Intensity:}}
Vision model research has largely focused on improving efficiency by reducing MACs.
Since accelerators greatly increase computational capabilities without proportionally increasing memory bandwidth, many models are bottlenecked by memory access and solely minimizing MACs will not improve performance.
Instead we must consider the Operational Intensity---the ratio of arithmetic operations to memory access.

\noindent{\textbf{MQA is efficient in hybrid models:}}
MHSA~\cite{vaswani2017attention} projects the queries, keys, and values into multiple spaces to capture different aspects of the information.
Multi-Query Attention (MQA)~\cite{shazeer2019fast} simplifies this by sharing keys and values across all heads. 
While large language models require multiple query heads, they can share a single head for keys and values without sacrificing accuracy~\cite{chowdhery2022palm}~\cite{lee2023platypus}.
When the number of batched tokens is small compared to the feature dimensions, sharing one head across keys and values reduces memory bandwidth requirements---significantly improving Operational Intensity.
In hybrid mobile vision models, the tokens are often small compared to features because attention is only used in the low-resolution later stages with high feature dimensions and because batch size one operation is common.
Our experiments confirm MQA's advantage in hybrid models. 
As shown in \cref{tab:attention_ablation} MQA achieves $>$39\% acceleration on EdgeTPUs and Samsung S23 GPU with negligible quality loss (-0.03\%) compared to MHSA.
MQA also reduces MACs and model parameters by $>$25\%. 
To our knowledge, we are the first to use MQA for mobile vision.
Furthermore, we introduce an additional Einsum optimization (see \cref{appendix:einsum}), specifically tailored for accelerated inference on hardware accelerators.
\begin{table}[t]
    \centering
    \scriptsize
    \caption{\textbf{MQA Impact:} MNv4-Conv-L base model. Attention blocks are added to the last stage. Percentage improvements only consider attention block latency \vs MHSA.}
    \begin{tabular}{ c c c c c c c}
        \toprule
        model & Top-1  & MACs & Params & \multicolumn{2}{c}{EdgeTPU}  & Samsung S23 \\
        & Acc(\%) &  (G) & (M) & Pixel 7 & Pixel 8  & GPU \\
        \midrule
        base model & 84.88 & 6.0 & 30.9 & 4.31 ms & 2.35 ms& 13.15 ms\\
        \midrule
        +3 MHSA & 85.27 & 6.7 & 36.0 & 9.69 ms & 2.76 ms & 16.46 ms\\
        \midrule
        +3 MQA  & 85.24 & 6.5  & 34.7  & 5.16 ms  & 2.60 ms & 15.10ms \\
        & (-0.03\%) & \textbf{(-28.6\%)} & \textbf{(-25.5\%)} & \textbf{(-84.2\%)} & \textbf{(-39.0\%)} & (\textbf{-41.1\%}) \\
        
        \bottomrule
    \end{tabular}
    \label{tab:attention_ablation}
\end{table}

\noindent\textbf{Incorporate asymmetric spatial down-sampling:}
Drawing inspiration from MQA, which utilizes asymmetric computation across queries, keys, and values, we add Spatial Reduction Attention (SRA)~\cite{wang2021pyramid} to our optimized MQA block to downscale key and value resolution while retaining high-resolution queries. 
This strategy is motivated by the observed correlation between spatially adjacent tokens in hybrid models attributed to spatial mixing convolution filters in early layers. 
Unlike~\cite{wang2021pyramid}, our method replaces AvgPooling with a stride-2 3x3 DW for spatial reduction---a cost-effective way to boost model capacity.

\noindent\textbf{Mobile MQA} Here we present our Mobile MQA block:
\begin{equation}
\begin{aligned}
\text{Mobile\_MQA}(\mathbf{X}) &= \text{Concat}(\text{attention}_1, \dots, \text{attention}_n)\mathbf{W}^O \\
\text{where} \ \text{attention}_j &= \text{softmax}\left(\frac{(\mathbf{X}\mathbf{W}^{Q_j})(SR(\mathbf{X})\mathbf{W}^K)^T}{\sqrt{d_k}}\right)(SR(\mathbf{X})\mathbf{W}^V)
\end{aligned}
\label{eq:mobile_mqa}
\end{equation}
where $SR$ denotes either spatial reduction, our stride-2 DW, or, if spatial reduction isn't used, the identity function. 
As shown in \cref{tab:ablation_downsampling}, asymmetric spatial down-sampling adds >20\% efficiency with minimal accuracy loss (-0.06\%).

\begin{table}[t]
\centering
\caption{\small \textbf{Impact of Downsampling on Mobile MQA:} MNv4-Hybrid-M base model on Samsung S23. Stride-2 down-sampling is applied at penultimate 16x16 stage.}
\label{tab:ablation_downsampling}
\scriptsize
\begin{tabular}{c c c c c}
\toprule
down-sampling on KV & Top-1 Acc & MACs (G) & CPU (ms) & GPU (ms) \\
\midrule
No & 80.77 & 1.285 & 15.8 & 7.4 \\
Yes & 80.71 & 1.245 & 12.8 & 5.9 \\
\midrule
Efficiency Gain & - & \textbf{+3\%} & \textbf{+23\%} & \textbf{+25\%} \\
\bottomrule
\end{tabular}
\end{table}

\section{Design of MNv4 Models}
\noindent\textbf{Our Design Philosophy: Simplicity Meets Efficiency.}
In developing the latest MobileNets, our core goal was Pareto optimality across diverse mobile platforms.
To achieve this, we started by conducting extensive correlation analyses on existing models and hardware. 
Through empirical examination, we found a set of components and parameters that ensure high correlations between cost models (the prediction of cost of latency) across various devices while approaching the Pareto frontier in performance.
Our investigation unveiled critical insights:

\textit{Multi-path efficiency concerns}: Group convolutions~\cite{shufflenet} and similar multi-path designs, despite lower MAC counts, can be less efficient due to memory access complexity.

\textit{Hardware support matters}: Advanced modules like Squeeze and Excite (SE)~\cite{hu2018squeeze}, GELU~\cite{hendrycks2016gaussian}, and LayerNorm~\cite{ba2016layer} are not well supported on DSPs, with LayerNorm also lagging behind BatchNorm~\cite{ioffe2015batch}, and SE is slow on accelerators.

\textit{The Power of Simplicity}:  Conventional components – depthwise and pointwise convolutions, ReLU~\cite{nair2010rectified}, BatchNorm, and simple attention (e.g., MHSA) –  demonstrate superior efficiency and hardware compatibility.

Based on these findings, we established a set of design principles:
\begin{itemize}
    \item \textbf{Standard Components}: We prioritize widely supported elements for seamless deployment and hardware efficiency.
    \item \textbf{Flexible UIB Blocks}: Our searchable UIB block lets NAS tune spatial and channel mixing, adjust receptive fields, and improve hardware utilization.
    \item \textbf{Employ Straightforward Attention}: Our Mobile MQA mechanism prioritizes simplicity for optimal performance.
\end{itemize}

These principles allow MobileNetV4 to be mostly  Pareto-optimal on all hardware evaluated.
In the following, we detail our refined NAS recipe for UIB model search, outline specific search configurations for various MNv4-Conv model sizes, and explain the construction of hybrid models.

\subsection{Refining NAS for Enhanced Architectures}
To effectively instantiate the UIB blocks, we adopt TuNAS~\cite{tunas} with tailored enhancements for improved performance.
We use use per-size searches and search spaces instead of using fixed scaling rules such as in EfficientNet\cite{tan2019efficientnet}.

\noindent\textbf{Enhanced Search Strategy}: Our approach mitigates TuNAS's bias towards smaller filters and expansion factors, attributed to parameter sharing, by implementing a two-stage search. 
This strategy addresses the variance in parameter counts between UIB's depthwise layers and other search options.

\textit{Coarse-Grained Search:} Initially, we focus on determining optimal filter sizes while maintaining fixed parameters: an inverted bottleneck block with a default expansion factor of 4 and a 3x3 depthwise kernel.

\textit{Fine-Grained Search:} Building on the initial search's outcomes, we search the configuration of UIB's two depthwise layers (including their presence and kernel size of either 3x3 or 5x5), keeping the expansion factor constant at 4.

\cref{tab:two_stage_search} demonstrates the enhanced efficiency and model quality achieved through our two-stage search compared to a conventional one-stage search, where a unified search space was explored in a single TuNAS pass.

\begin{table}[t]
    \centering
    \scriptsize
    \caption{Comparison between one-stage and two-stage searches, highlighting accuracy improvements and latency reduction on Pixel 6 EdgeTPU.}
    \begin{tabular}{l c c c}
        \hline
        Search Method & Top-1 Acc (Val) & Top-1 Acc (Train) & Pixel 6 EdgeTPU (ms) \\
        \hline
        One-stage & 81.26 & 74.64 &  3.85 \\
        Two-stage & 81.48 \textbf{(+0.22)} & 78.24 \textbf{(+3.60)} & 3.67 \textbf{(-4.68\%)} \\
        \hline
    \end{tabular}
\label{tab:two_stage_search}
\end{table}

\noindent\textbf{Enhancing TuNAS with Robust Training:}
The success of TuNAS hinges on accurately evaluating architecture quality, crucial for reward calculation and policy learning. 
Originally, TuNAS leveraged ImageNet-1k for training the SuperNet, but ImageNet performance is notably affected by data augmentation, regularization, and hyper-parameter choices. 
Given TuNAS's evolving architecture samples, finding a stable set of hyper-parameters is challenging.

We address this with an offline distillation dataset, eliminating the need for extra augmentations and reducing sensitivity to regularization and optimization settings. 
The JFT distillation dataset, as detailed in \cref{sec:distill}, serves as our training set for TuNAS, with notable improvements shown in \cref{tab:search_with_jft}.
Acknowledging that depth-scaling surpasses width-scaling in extended training sessions~\cite{revisiting_resnet}, we extend TuNAS training to 750 epochs, yielding deeper, higher-quality models.

\begin{table}[t!]
    \centering
    \scriptsize
\caption{\small Performance Boost from JFT Distillation: NAS Training on ImageNet-1k vs. JFT Data. Highlights efficiency improvements and slight accuracy differences.}
        \begin{tabular}{l c c c c c }
            \toprule
NAS Dataset & Top-1 Acc (Val/Train) & MACs & Params & Pixel 4 GPU & Pixel 6 CPU \\
            \midrule
ImageNet & 82.4 / 72.9 & 7.2G & 43.5M & 59.2ms & 70.4ms \\
JFT distill & 82.3 / 74.0 & 6.2G & 34.4M & 51.0ms & 67.3ms \\
\midrule
Gain & -0.1 / +1.1 & \textbf{+13.9\%} & \textbf{+20.9\%} & \textbf{+13.9\%} & \textbf{+4.4\%} \\
\bottomrule
        \end{tabular}
\label{tab:search_with_jft}
\end{table}

\subsection{Optimization of MNv4 Models}
We constructed MNv4-Conv models from NAS-optimized UIB blocks, tailoring them for specific resource constraints.
More details are given in \cref{appendix:search_space}.
In line with other hybrid models, we found that adding attention to the last stages of convolution models is most effective. 
In MNv4-Hybrid models, we interlace Mobile MQA blocks with UIB blocks for enhanced performance.
For comprehensive model specifications, refer to \cref{appendix:model_detail}.

\section{Results}
\label{sec:experiments}
In this section, we demonstrate the mostly Pareto-optimal performance of MobileNetV4 (MNv4) on ImageNet-1K classification and COCO object detection.

\subsection{ImageNet classification}
\label{subsec:imagenet}
\noindent \textbf{Experimental Setup}:
To assess model architecture performance, we train exclusively with the ImageNet-1k~\cite{imagenet} training split and measure Top-1 accuracy on its validation split. 
Our latency analysis includes a representative selection of mobile hardware, including ARM Cortex CPUs (Pixel 6, Samsung S23), Qualcomm Hexagon DSP (Pixel 4), ARM Mali GPU (Pixel 7), Qualcomm Snapdragon (S23 GPU), Apple Neural Engine, and Google EdgeTPU. Our complete training recipe is detailed in the \cref{appendix:training_detail}.

We benchmark our models against the leading efficient models, including hybrid (MiT-EfficientViT~\cite{mit-efficientvit}, FastViT~\cite{vasu2023fastvit}, NextViT~\cite{li2022nextvit}) and convolutional models (MobileOne~\cite{vasu2023mobileone}, ConvNext~\cite{convnext}, and previous MobileNets ~\cite{mobilenet}~\cite{mobilenet_v2}~\cite{mobilenet_v3}) based on their reported Top-1 Accuracies and our latency evaluations. 
We used modern training recipes to improve MobileNetV1-V3 accuracy: a +3.4\% (to 74.0\%) for V1, +1.4\% (to 73.4\%) for V2, and +0.3\% (to 75.5\%) for V3. 
These new figures are used throughout the paper to isolate architectural advancements.

\begin{table}[t]
\small
\caption{
\small \textbf{Classification results on ImageNet-1K~\cite{imagenet}, along with on-device benchmarks.} Median latency is reported. A $-$ indicates that we did not benchmark a model due to missing corresponding model file for a platform. \textit{Failed} indicates that the model is not supported by the platform. Dividers denote approximate latency classes.
}
\resizebox{\textwidth}{!}{
\begin{tabular}{l c c c c c c c c c c}
\toprule
 & & & & \multicolumn{7}{c}{Latency (ms)} \\
 & &  Params & MACs & Pixel 6 & Pixel 8  & iPhone 13 & Pixel 4 & Pixel 7 & \multicolumn{2}{c}{Samsung S23} \\
Model & Top-1 & (M) & (G) &  CPU & EdgeTPU & CoreML & Hexagon & GPU & CPU & GPU \\
\midrule
MobileNet-V2-0.5x~\cite{mobilenet_v2} & 66 & 2.0 & 0.1 & 2.4 & 0.7 & 0.5& 2.9 & 8.3 & 1.8 & 1.9 \\
MobileNet-V3L-0.5x~\cite{mobilenet_v3} & 69.2 & 2.7 & 0.1 & 2.4 & 0.8 & 0.45 & 3.5 & 9.9 & 2.0 & 2.1 \\
MobileOne-S0~\cite{vasu2023mobileone}  & 71.4 & 2.1 & 0.3 & 4.2 & 0.7 & 0.5 & 2.9 & 10.7 & 3.3 & 1.7 \\
MobileNet-V2~\cite{mobilenet_v2} & 73.4 & 3.5 & 0.3 & 5.0 & 0.7 & 0.7 & 3.9 & 13.6 & 4.1 & 2.5 \\
\textbf{MNv4-Conv-S} & \textbf{73.8} & \textbf{3.8} & \textbf{0.2} & \textbf{2.4} & \textbf{0.7} & \textbf{0.6} & \textbf{2.4} & \textbf{8.4} & \textbf{1.8} & \textbf{2.0} \\
MobileNet-V1~\cite{howard2017mobilenets} & 74.0 & 4.2 & 0.6 & 6.1 & 0.8 & 0.7 & 3.2 & 13.0 & 4.6 & 2.1 \\
\midrule
FastViT-T8$^{\dag}$~\cite{vasu2023fastvit} & 75.6 & 3.6 & 0.7 & 49.3 & 1.3 & 0.7 & \textit{Failed} & 40.7 & 43.6 & 24.7 \\
MobileNet-V2-1.5x~\cite{mobilenet_v2} & 76.8 & 6.8 & 0.7 & 9.3 & 0.9 & 1.0 & 5.6 & 16.4 & 7.3 & 3.3 \\
MultiHardware-MAX-1.5x~\cite{mobilenet_multi} & 77.9 & 8.9 & 0.8 & 9.8 & 1.0 & - & 5.7 & 23.2 & - & 4.1 \\
MultiHardware-AVG-1.5x~\cite{mobilenet_multi} & 78.2 & 10.0 & 1.0 & 12.0 & 1.1 & - & 6.1 & 20.3 & - & 4.5 \\
MobileNet-V2-2.0x~\cite{mobilenet_v2} & 78.4 & 11.2 & 1.1 & 13.9 & 1.1 & 1.5 & 6.9 & 19.1 & 10.6 & 4.2 \\
MobileOne-S4~\cite{vasu2023mobileone} & 79.4 & 14.8 & 1.5 & 26.7 & 1.7 & 1.5 & 9.0 & 28.6 & 19.4 & 5.9 \\
FastViT-S12$^{\dag}$~\cite{vasu2023fastvit} & 79.8 & 8.8 & 1.8 & 83.0 & 1.8 & 1.6 & \textit{Failed} & 75.0 & 69.2 & 47.0 \\
MIT-EfficientViT-B1-r224~\cite{mit-efficientvit} & 79.4 & 9.1 & 0.5 & - & - & 2.4 & - & - & 18.1 & 5.0 \\
\textbf{MNv4-Conv-M} & \textbf{79.9} & \textbf{9.2} & \textbf{1.0} & \textbf{11.4} & \textbf{1.1} & \textbf{1.1} & \textbf{7.3} & \textbf{18.1}  & \textbf{8.6} & \textbf{4.1} \\
\midrule
FastViT-SA12~\cite{vasu2023fastvit}  & 80.6 & 10.9 & 1.9 & 86.5 & 2.0 & 1.6 & \textit{Failed} & 79.6 & 69.5 & 52.1 \\
\textbf{MNv4-Hybrid-M} & \textbf{80.7} & \textbf{10.5} & \textbf{1.2} & \textbf{14.3} & \textbf{1.5} & - & \textbf{\textit{Failed}} & \textbf{17.9} & \textbf{10.8} & \textbf{5.9} \\
\midrule
FastViT-SA24~\cite{vasu2023fastvit} & 82.6 & 20.6 & 3.8 & 171.6 & 3.2 & 2.4 & \textit{Failed} & 131.9 & 136.3 & 107.5 \\
MIT-EfficientViT-B2-r256~\cite{mit-efficientvit} & 82.7 & 24.0 & 2.1 & - & - & 5.4 & - & - & 64.9 & 9.5 \\
\textbf{MNv4-Conv-L} & \textbf{82.9} & \textbf{31} & \textbf{5.9} & \textbf{59.9} & \textbf{2.4} & \textbf{3.0} & \textbf{20.8} & \textbf{37.6} & \textbf{43.0} & \textbf{13.2} \\
\midrule
ConvNext-S~\cite{convnext} & 83.1 & 50 & 8.7 & 314.9 & 3.7 & - & \textit{Failed} & 45.2 & 243.9 & 18.5 \\
NextViT-B~\cite{li2022nextvit} & 83.2 & 44.8 & 8.3 & - & - & - & - & - & - & - \\
\textbf{MNv4-Hybrid-L} & \textbf{83.4} & \textbf{35.9} & \textbf{7.2} & \textbf{87.6} & \textbf{3.8} & - & \textbf{\textit{Failed}} & \textbf{61.3} & \textbf{61.8} & \textbf{18.1} \\
MIT-EfficientViT-B3-r224~\cite{mit-efficientvit} & 83.5 & 49.0 & 4.0 & - & - & 12.2 & - & - & 125.9 & 18.4 \\
FastViT-SA36~\cite{vasu2023fastvit} & 83.6 & 30.4 & 5.6 & 241.6 & 4.3 & - & \textit{Failed} & 186.5 & 206.3 & 138.1 \\
\bottomrule
\end{tabular}}
\label{tab:classification_results}
\end{table}

\noindent\textbf{Results: } Our results, seen in \cref{fig:multi_hardware_pareto} and \cref{tab:classification_results}, demonstrate that MNv4 models are mostly Pareto-optimal across a range of accuracies and mobile targets, including CPUs, DSPs, GPUs, and accelerators like the Apple Neural Engine and Google EdgeTPU.

MNv4 performs notably well on CPU---roughly 2x faster than MobileNetV3 and substantially faster than iso-accuracy models.
On EdgeTPUs, MNv4 models are as accurate as MobileNetV3 and 2x as fast.
MNv4-Conv-M is >50\% faster than MobileOne-S4 and FastViT-S12 and has +1.5\% more Top-1 accuracy than MobileNetV2 at comparable latency.
On S23 GPU and iPhone 13 CoreML (ANE), MNv4 is mostly at the Pareto front. MIT-EfficientViT---which has the closest performance on S23 GPU---has >2x the latency as MNv4 on CoreML at the same accuracy. 
FastViT---optimized for Apple Neural Engine---is 2nd on CoreML but has >5x the latency of MNv4 on S23 GPU.
While some models, such as EfficientViT, reach the same accuracy with fewer MACs, MobileNetV4 models are optimized for high accuracy and minimal latency on the most hardware possible. 
Increasing MACs often decreases memory bandwidth and op complexity which is often more important for achieving this goal. 
Like many hybrid models, MNv4-hybrid models are not compatible with DSPs. 
MNv4-Conv models remain the top performers on DSP, emphasizing the compatibility and efficiency across diverse hardware provided by our UIB block, NAS recipe, and search spaces.
MNv4-Hybrid performs well on CPUs and accelerators which demonstrates the broad efficiency of Mobile MQA.

Mobile models should perform well on diverse hardware, but we show that many models fail to meet this requirement.
MobileNetV3 performs well on CPUs but not on EdgeTPU, DSPs, and GPUs. 
FastViT performs well on ANE but not on CPUs and GPUs. EfficientViT has good performance on GPUs but not on ANE.
In contrast, MNv4-Conv models achieves mostly-Pareto-optimal performance across CPUs, GPUs, DSPs, the Apple Neural Engine, and Google EdgeTPUs. 
This versatility ensures MNv4-Conv models can be easily deployed across the mobile ecosystem and sets a new benchmark for mobile model universality.

\subsection{COCO Object Detection}
\label{subsec:detection}

\textbf{Experimental Setup: }
We evaluate the effectiveness of MNv4 backbones for object detection tasks on the COCO~17~\cite{coco_2014} dataset.
We compare MNv4 medium backbones against SOTA backbones with a MAC count.
For each backbone, we build a detector using the RetinaNet~\cite{retinanet} framework.
We attach a $256$-d FPN~\cite{fpn} decoder to the P$3$ - P$7$ endpoints, as well as a $256$-d prediction head with $4$ convolutional layers.
As usual for mobile detectors, we use depth-separable convolutions for an efficient FPN decoder and box prediction head.
We train all models on COCO~17~\cite{coco_2014} for $600$-epochs.
Images are resized to $384px$ and augmented with random horizontal flip, random scale, and Randaug~\cite{randaugment}.
We exclude Shear and Rotate from Randaug, as those deteriorate small-object detection AP.
The models are trained with a $2048$ batch size, Adam~\cite{Adam}, and a $0.00003$ L2 weight decay, plus a cosine LR schedule with $24$ epochs warm-up.
The learning rate is tuned per-model.
For all baselines, filter multipliers are tuned to similar MACs. 
Following classification, MobileNetV4 backbones are trained using a $0.2$ stochastic drop~\cite{stochasticDepth}.
MobileNet baselines were from Tensorflow Model Garden~\cite{tensorflow_model_garden} implementation.
EfficientFormer was reimplemented in Tensorflow.

\noindent{\textbf{Results: }}
Results are reported in \cref{tab:coco_17}. Parameters, MACs and benchmarks are computed using the entire detector at the $384px$ input resolution. 
The MNv4-Conv-M detector achieves 32.6\% AP, similar to MobileNetMultiAvg and MobileNetV2. 
However, this model is $12$\% faster than MobileNetMultiAvg and $23$\% faster than MobileNetV2 on Pixel 6 CPU. 
The MNv4-Hybrid-M detector gets $+1.6$\% AP over MNv4-Conv-M while running $18$\% slower on Pixel 6 CPU.
This demonstrates the effectiveness of MNv4 hybrid models on tasks like object detection.

\begin{table}[t]
\centering
\scriptsize
\caption{Object detection results on the COCO-17~\cite{coco_2014} Val. set. The width-multiplier is reported next to the MobileNet backbones that were scaled-up.}
\begin{tabular}{lcccc}
\toprule
&  COCO & MACs & Params & Pixel 6 CPU \\
Backbone & Val AP & (G) & (M) & latency (ms) \\
\midrule
EfficientFormer L1~\cite{efficientformer} & 29.5 & 6.54 & 12.77& 84.3  \\
MobileNet v1 @ 1.5~\cite{mobilenet} &  31.0 & 6.68 & 9.05 & 66.4 \\
\bf{MNv4-Conv-M} & \bf{32.6} & \bf{5.06} & \bf{9.79} & \bf{51.3} \\
\midrule
MobileNet Multi-AVG @ 1.5~\cite{mobilenet_multi} & 32.7 & 5.42 & 9.51 & 58.1 \\
MobileNet v2 @ 2.0~\cite{mobilenet_v2} &  32.9 & 5.81 & 10.15 & 66.4 \\
MobileNet v3 Large~\cite{mobilenet_v3} @ 2.0 & 33.2 & 4.99 & 17.92 & 59.9 \\
\bf{MNv4-Hybrid-M} & \bf{34.0} & \bf{5.62} & \bf{11.15} & \bf{60.5} \\
\bottomrule
\end{tabular}
\label{tab:coco_17}
\end{table}

\section{Enhanced distillation recipe}
\label{sec:distill}
Complementing architectural innovation, distillation is a powerful tool for enhancing machine learning efficiency. 
This is particularly true for mobile models where distillation can greatly increase accuracy without increasing latency.
Building upon the Patient Teacher distillation baseline~\cite{beyer2022knowledge}, we introduce two novel techniques to further boost performance.

\noindent\textbf{Dynamic Dataset Mixing:}
Data augmentation is crucial for distillation performance.
While prior methods rely on a fixed augmentation sequence, we find that dynamically mixing multiple datasets with diverse augmentation strategies improves distillation. Our experiments use three distillation datasets:
\begin{description}
    \item[$\mathcal{D}_1$]: 
    Inception Crop~\cite{inception} followed by RandAugment~\cite{cubuk2020randaugment} l2m9 applied to 500 ImageNet-1k replicas.
    \item[$\mathcal{D}_2$]: Inception Crop followed by extreme Mixup~\cite{zhang2017mixup} applied to 1000 ImageNet-1k replicas (mirroring the Patient Teacher approach).
    \item[$\mathcal{D}_1$ + $\mathcal{D}_2$]: 
  A dynamic mixture of {$\mathcal{D}_1$} and {$\mathcal{D}_2$} during training.
\end{description}

Our results (\cref{tab:student_result}) show that {$\mathcal{D}_2$} outperforms {$\mathcal{D}_1$} (84.1\% vs. 83.8\% student accuracy), but a dynamic mixture of the two ({$\mathcal{D}_1$ + $\mathcal{D}_2$}) elevates accuracy to 84.4\% (+0.3\%).
This suggests that mixing expands the augmented image space, increases difficulty and diversity, and leads to improved student performance.

\noindent\textbf{JFT Data Augmentation: }
To increase training data volume, we add in-domain, class-balanced data by resampling the JFT-300M~\cite{sun2017revisiting} dataset to 130K images per class (130M total). 
Following Noisy Student\cite{xie2020self} and using EfficientNet-B0 trained on ImageNet-1K, we select images with a relevance threshold above 0.3.
For classes with abundant data, we choose the top 130K images; for rare classes, we replicate images for balance.
This dataset is replicated 10x. 
Due to JFT's complexity, we apply weaker augmentations (Inception Crop + RandAugment l2m5).
This is dataset $\mathcal{D}_3$.
\cref{tab:student_result} shows that using solely $\mathcal{D}_3$ drops accuracy by 2\%.
However, combining ImageNet and JFT data ({$\mathcal{D}_1$ + $\mathcal{D}_2$} + $\mathcal{D}_3$) raises accuracy by +0.6\%. The additional data improves generalization.

\noindent\textbf{Our distillation recipe:}
Our combined distillation recipe dynamically mixes datasets 
$\mathcal{D}_1$, $\mathcal{D}_2$, and $\mathcal{D}_3$ for diverse augmentations and leverages class-balanced JFT data. 
As shown in \cref{tab:student_result} and \cref{tab:distillation_effectiveness}, our method improves top-1 accuracy $>$0.8\% over the previous SOTA~\cite{beyer2022knowledge}. 
Training an MNv4-Conv-L student model for 2000 epochs yields 85.9\% top-1 accuracy. 
Our approach is effective: the student has 15x fewer parameters and 48x fewer MACs than its teacher (EfficientNet-L2), but is only 1.6\% less accurate.
MNv4-Conv-Hybrid reaches 87.0\% top-1 accuracy by combining this distillation with pretraining on JFT.
More details of our distillation recipe can be found in \Cref{appendix:distillation}.

\begin{table}[t]
    \centering
        \centering
        \scriptsize
        \caption{\small Distillation results using MNv4-Conv-L as student, highlighting gains over SOTA and marking our contributions explicitly.}
        \begin{tabular}{cccccc}
            \toprule
          Dataset & Data source & Augmentations & Mixing & \multicolumn{2}{c}{Top-1 Acc (Val/Train)} \\
            \cline{5-6}
         & & & Ratio & 400 epochs & 2000 epochs \\
            \midrule
           $\mathcal{D}_1$ & 1000$\times$ & Inception Crop \& & - & 83.8/86.6 & -\\
           & ImageNet-1k & RandAug l2m9 & & & \\
            \midrule
            $\mathcal{D}_2$  & 1000$\times$ & Inception Crop \&  & - & 84.1/85.6 & -\\
            (SOTA~\cite{beyer2022knowledge}) & ImageNet-1k & Extreme Mixup & \\
            \midrule
            $\mathcal{D}_3$  & 10$\times$  & Inception Crop \& & - & 81.8/84.1 & -\\
           & JFT subset & RandAug l2m5 & & & \\
            \midrule
            \textbf{Ours}: \textbf{$\mathcal{D}_1$ + $\mathcal{D}_2$} &  & & 1:1 & \textbf{84.4}/85.0 ($+$\textbf{0.3})& - \\
            \midrule
            \textbf{Ours}: \textbf{$\mathcal{D}_2$ + $\mathcal{D}_3$} &  & & 1:1 & \textbf{84.7}/82.7 ($+$\textbf{0.6}) & - \\
            \midrule      
            \textbf{Ours}: \textbf{$\mathcal{D}_1$ + $\mathcal{D}_2$ + $\mathcal{D}_3$} &  & & 1:1:2 & \textbf{84.9}/82.6 ($+$\textbf{0.8})  & 85.9/85.5 ($+$\textbf{1.8})\\
            \bottomrule  
        \end{tabular}
        \label{tab:student_result}
\end{table}

\begin{table}
\centering
\scriptsize
\caption{\textbf{Top-1 Accuracy Comparison Across Training Approaches}: This table contrasts baseline ImageNet-1k training, SOTA distillation, and our distillation.
}
\begin{tabular}{c | c | c | c | c}
 \toprule
 Model & IN-1k Only & SOTA & \textbf{Our} & \textbf{Our Gain} Over \\
  & Only & Distill\cite{beyer2022knowledge} & \textbf{Distill} & IN-1k / SOTA\\
 \midrule
 MNv4-Conv-S  & 73.8 & - & \textbf{75.5} & \textbf{+1.7} / - \\
 MNv4-Conv-M & 79.9 & 81.5 & \textbf{82.7} & \textbf{+2.8} / \textbf{+1.2} \\
 MNv4-Hybrid-M & 80.7 & 82.7 & \textbf{83.7} & \textbf{+3.0} / \textbf{+1.0} \\
 MNv4-Conv-L & 82.9 & 84.4 & \textbf{85.9} & \textbf{+3.0} / \textbf{+1.5} \\
 MNv4-Hybrid-L & 83.4 & 85.7 & \textbf{86.6} & \textbf{+3.2} / \textbf{+0.9} \\
 \bottomrule
\end{tabular}
\label{tab:distillation_effectiveness}
\end{table}

\section{Conclusion}
In this paper, we presented MobileNetV4, a series of universal, efficient models that run efficiently across the mobile ecosystem. Multiple advances make MobileNetV4 mostly-Pareto-optimal on all mobile CPUs, GPUs, DSPs and specialized accelerators, a characteristic not found in any other models tested. We introduced the Universal Inverted Bottleneck and Mobile MQA layers and combined them with improved NAS recipes. With these and a novel, SOTA distillation approach, we achieve 87\% ImageNet-1K accuracy at 3.8ms Pixel 8 EdgeTPU latency, setting a new state-of-the-art. Finally, we introduced a framework for understanding model universality on heterogeneous devices. We hope the novel contributions and analysis further spur advances in mobile computer vision.

\section*{Acknowledgements}
We appreciate Tammo Spalink, Yeqing Li, Sage Stevens, Bob Muniz, Liviu Panait, David Wood, Lynn Nguyen, and Lucas Beyer for their support while developing this work.
We also thank the MLPerf Mobile working group for their feedback and collaboration.\footnote{MobileNetV4-Conv-L was used for v4.0 of {\scriptsize\url{https://mlcommons.org/benchmarks/inference-mobile/}}}
We particularly thank Ross Wightman for his reimplementation of MobileNetV4 for \texttt{timm} and training recipe improvements.\footnote{MobileNetV4 models are available in \texttt{timm} at {\scriptsize \url{https://huggingface.co/collections/timm/mobilenetv4-pretrained-weights-6669c22cda4db4244def9637}}}

{\small
\bibliographystyle{ieee_fullname}
\bibliography{main}
}
\clearpage

\appendix

\section{Search space details}
\label{appendix:search_space}
The following is how we construct the search space for NAS.

\noindent\textbf{Search Space Construction:}
\begin{itemize}
    \item \textit{Fixed Initial Layers:} We started with a Conv2D layer (3x3 kernel, stride 2) in the first stage for quick resolution reduction, followed by NAS-optimized FusedIB blocks (stride 2) in the second stage to balance between efficiency and accuracy.
    \item \textit{NAS-Driven Optimization:} The NAS process precisely determined the ideal number of UIB blocks and parameter instantiations across the remaining four stages, ensuring an optimal structure for performance.
    \item \textit{Fixed Head Layers}: We use the same head layer configuration as MobileNet V3.
\end{itemize}

Observing that pointwise convolutions within UIB blocks tend to exhibit low operational intensity at higher resolutions,
we prioritized operations with higher computational density in the initial layers to balance efficiency and accuracy.

\noindent\textbf{Our optimization targets:}
\vspace{-2.5mm}
\begin{itemize}
    \item MNv4-Conv-S: Dual goals---285M MACs and 0.2ms latency (Pixel 6 EdgeTPU, 224px inputs). 
    \item MNv4-Conv-M: 0.6ms latency (Pixel 6 EdgeTPU, 256px inputs).
    \item MNv4-Conv-L: Dual latency goals of 2.3ms (Pixel 6 EdgeTPU) and 2.0ms (Pixel 7 EdgeTPU) with 384px inputs.
\end{itemize}
\vspace{-2.5mm}
To be noted, by restricting our search space to components with well-correlated cost models across devices, we found that EdgeTPU latency optimization directly yields universally efficient models, as demonstrated in later sections.

\section{Benchmarking methodology}
\label{appendix:benchmarking}
We applied a consistent benchmarking strategy across various mobile platforms, with an exception for the Apple Neural Engine.
To enhance efficiency, models were converted to TensorFlow Lite format and quantized to INT8 for mobile CPUs, Hexagon, and EdgeTPUs, while FP16 was used for mobile GPUs.
We run each model roughly 1000 times and take the mean latency of those runs. We then repeat that process 5 times for each model and report the median of means.
To optimize performance, we set the CPU affinity to the fastest core and use the XNNPACK backend for CPU evaluations.
In contrast, for  benchmarks on the Apple Neural Engine (conducted on an iPhone 13 with iOS 16.6.1, CoreMLTools 7.1, and Xcode 15.0.1 for profiling), PyTorch models were converted to CoreML's MLProgram format in Float16 precision, with float16 MultiArray inputs to minimize input copying.

\section{Training setup for ImageNet-1k classification}
\label{appendix:training_detail}

To enhance model performance, our training recipe incorporates widely adopted data augmentation techniques and regularization methods.
For data augmentation, we use Inception Crop~\cite{inception}, horizontal flip, RandAugment~\cite{randaugment}, Mixup~\cite{zhang2017mixup}, and CutMix~\cite{yun2019cutmix}. 
For regularization, we apply L2 normalization and stochastic depth drop~\cite{stochasticDepth}.
The intensity of augmentation and regularization is adjusted according to model size, as detailed in \cref{tab:imagenet_setting}.

\begin{table}
\centering
\scriptsize
\caption{Training hyper-parameters for ImageNet-1k classification.}
\begin{tabular}{lccccc}
\toprule
 & Conv-S & Conv-M & Hybrid-M & Conv-L & Hybrid-L  \\
\midrule
Batch size & 4096 & 4096 & 16384 & 16384 & 16384  \\
Peak learning rate & 0.002 & 0.004 & 0.016 & 0.004 & 0.01 \\
Cosine decay alpha & 0.0   & 0.0   & 0.0   & 0.0   & 0.001 \\
Cosine decay epochs & 9600 & 500 & 500 & 500 & 500 \\
Warm-up epochs & 5 & 5 & 20 & 20 & 20  \\
Training epochs & 9600 & 500 & 500 & 500 & 500  \\
\midrule
AdamW weight decay & 0.01 & 0.1 & 0.1 & 0.2 & 0.2 \\
AdamW $\beta_1$ & 0.6 & 0.9 & 0.9 & 0.9 & 0.9 \\
AdamW $\beta_2$ & 0.999 & 0.999 & 0.999 & 0.999 & 0.999 \\
AdamW $\epsilon$ & $10^{-6}$ & $10^{-7}$ & $10^{-7}$ & $10^{-7}$ & $10^{-7}$  \\
EMA decay & 0.9999 & - & - & - & -  \\
\midrule
L2-regularization & $10^{-5}$ & - & - & - & - \\
Gradient clipping & - & - & - & - & - \\
Label smoothing & 0.1 & 0.1 & 0.1 & 0.1 & 0.1 \\
Dropout & 0.3 & 0.2 & 0.2 & 0.2 & 0.2  \\
Peak Stochastic Depth drop rate & 0 & 0.075 & 0.075 & 0.35 & 0.35  \\
\midrule
RandAugment probability & 0.5 & 0.7 & 0.7 & 1.0 & 1.0 \\
RandAugment layers & 2 & 2 & 2 & 2 & 2 \\
RandAugment magnitude & 9 & 15 & 15 & 15 & 15  \\
RandAugment excluded ops & Cutout & Cutout & Cutout & Cutout & Cutout \\
Mixup/Cutmix probability & - & - & - & 0.3 & 0.3 \\
Mixup $\alpha$ & - & - & - & 0.8 & 0.8  \\
Cutmix $\alpha$ & - & - & - & 1.0 & 1.0  \\
Mixup/Cutmix switch probability & - & - & - & 0.5 & 0.5  \\
\bottomrule
\end{tabular}
\label{tab:imagenet_setting}
\end{table}

\section{Model details}
\label{appendix:model_detail}
The architecture details of our MNv4 models are described from
\cref{arch_0p5x} to \cref{arch_8x_max}.

Now, let's examine the details of the TuNAS-optimized MNv4-Conv models.
TuNAS optimized macro architecture strategically combines four UIB instantiations: Extra DW, ConvNext, IB, and FFN. 
This combination demonstrates the flexibility of UIB and the importance of using different instantiation blocks in different stages of the network.
Specifically, at the start of each searchable stage, where the spatial resolution significantly drops, ExtraDW emerges as the preferred choice. 
The design of duo depthwise layers in ExtraDW helps to enlarge the receptive field, enhances spatial mixing, and effectively mitigates resolution loss. 
Similarly, ExtraDW is frequently selected in the early stages of MNv4-Conv models for similar reasons.
For the final layers, where preceding layers have conducted substantial spatial mixing, FFN and ConvNext are chosen because channel mixing provides a larger incremental gain.

\begin{table}
\centering
\scriptsize
\caption{Architecture specification of MNv4-Conv-S.}
\begin{tabular}{c|c|c|c|c|c|c}
\toprule[0.2em]
Input & Block & DW $K_1$ & DW $K_2$ & Expanded Dim & Output Dim & Stride \\
\midrule[0.2em]
$224^2\times3$ & Conv2D & - & $3\times3$ & - & 32 & 2 \\
\midrule[0.1em]
$112^2\times32$ & FusedIB & - & $3\times3$ & 32 & 32 & 2 \\
\midrule[0.1em]
$56^2\times32$ & FusedIB & - & $3\times3$ & 96 & 64 & 2 \\
\midrule[0.1em]
$28^2\times64$ & ExtraDW & $5\times5$ & $5\times5$ & 192 & 96 & 2 \\
$14^2\times96$ & IB & - & $3\times3$ & 192 & 96 & 1 \\
$14^2\times96$ & IB & - & $3\times3$ & 192 & 96 & 1 \\
$14^2\times96$ & IB & - & $3\times3$ & 192 & 96 & 1 \\
$14^2\times96$ & IB & - & $3\times3$ & 192 & 96 & 1 \\
$14^2\times96$ & ConvNext & $3\times3$ & - & 384 & 96 & 1 \\
\midrule[0.1em]
$14^2\times96$ & ExtraDW & $3\times3$ & $3\times3$ & 576 & 128 & 2 \\
$7^2\times128$ & ExtraDW & $5\times5$ & $5\times5$ & 512 & 128 & 1 \\
$7^2\times128$ & IB & - & $5\times5$ & 512 & 128 & 1 \\
$7^2\times128$ & IB & - & $5\times5$ & 384 & 128 & 1 \\
$7^2\times128$ & IB & - & $3\times3$ & 512 & 128 & 1 \\
$7^2\times128$ & IB & - & $3\times3$ & 512 & 128 & 1 \\
\midrule[0.1em]
$7^2\times128$ & Conv2D & - & $1\times1$ & - & 960 & 1 \\
$7^2\times960$ & AvgPool & - & $7\times7$ & - & 960 & 1 \\
$1^2\times960$ & Conv2D & - & $1\times1$ & - & 1280 & 1 \\
$1^2\times1280$ & Conv2D & - & $1\times1$ & - & 1000 & 1 \\
\bottomrule[0.2em]
\end{tabular}
\label{arch_0p5x}
\end{table}
\begin{table}
\centering
\scriptsize
\caption{Architecture specification of MNv4-Conv-M.}
\begin{tabular}{c|c|c|c|c|c|c}
\toprule[0.2em]
Input & Block & DW $K_1$ & DW $K_2$ & Expanded Dim & Output Dim & Stride \\
\midrule[0.2em]
$256^2\times3$ & Conv2D & - & $3\times3$ & - & 32 & 2 \\
\midrule[0.1em]
$128^2\times32$ & FusedIB & - & $3\times3$ & 128 & 48 & 2 \\
\midrule[0.1em]
$64^2\times48$ & ExtraDW & $3\times3$ & $5\times5$ & 192 & 80 & 2 \\
$32^2\times80$ & ExtraDW & $3\times3$ & $3\times3$ & 160 & 80 & 1 \\
\midrule[0.1em]
$32^2\times80$ & ExtraDW & $3\times3$ & $5\times5$ & 480 & 160 & 2 \\
$16^2\times160$ & ExtraDW & $3\times3$ & $3\times3$ & 640 & 160 & 1 \\
$16^2\times160$ & ExtraDW & $3\times3$ & $3\times3$ & 640 & 160 & 1 \\
$16^2\times160$ & ExtraDW & $3\times3$ & $5\times5$ & 640 & 160 & 1 \\
$16^2\times160$ & ExtraDW & $3\times3$ & $3\times3$ & 640 & 160 & 1 \\
$16^2\times160$ & ConvNext & $3\times3$ & - & 640 & 160 & 1 \\
$16^2\times160$ & FFN & - & - & 320 & 160 & 1 \\
$16^2\times160$ & ConvNext & $3\times3$ & - & 640 & 160 & 1 \\
\midrule[0.1em]
$16^2\times160$ & ExtraDW & $5\times5$ & $5\times5$ & 960 & 256 & 2 \\
$8^2\times256$ & ExtraDW & $5\times5$ & $5\times5$ & 1024 & 256 & 1 \\
$8^2\times256$ & ExtraDW & $3\times3$ & $5\times5$ & 1024 & 256 & 1 \\
$8^2\times256$ & ExtraDW & $3\times3$ & $5\times5$ & 1024 & 256 & 1 \\
$8^2\times256$ & FFN & - & - & 1024 & 256 & 1 \\
$8^2\times256$ & ConvNext & $3\times3$ & - & 1024 & 256 & 1 \\
$8^2\times256$ & ExtraDW & $3\times3$ & $5\times5$ & 512 & 256 & 1 \\
$8^2\times256$ & ExtraDW & $5\times5$ & $5\times5$ & 1024 & 256 & 1 \\
$8^2\times256$ & FFN & - & - & 1024 & 256 & 1 \\
$8^2\times256$ & FFN & - & - & 1024 & 256 & 1 \\
$8^2\times256$ & ConvNext & $5\times5$ & - & 512 & 256 & 1 \\
\midrule[0.1em]
$8^2\times256$ & Conv2D & - & $1\times1$ & - & 960 & 1 \\
$8^2\times960$ & AvgPool & - & $8\times8$ & - & 960 & 1 \\
$1^2\times960$ & Conv2D & - & $1\times1$ & - & 1280 & 1 \\
$1^2\times1280$ & Conv2D & - & $1\times1$ & - & 1000 & 1 \\
\bottomrule[0.2em]
\end{tabular}
\label{arch_2x}
\end{table}
\begin{table}
\scriptsize
\centering
\caption{Architecture specification of MNv4-Hybrid-M.}
\begin{tabular}{c|c|c|c|c|c|c}
\toprule[0.2em]
Input & Block & DW $K_1$ & DW $K_2$ & Expanded Dim & Output Dim & Stride \\
\midrule[0.2em]
$256^2\times3$ & Conv2D & - & $3\times3$ & - & 32 & 2 \\
\midrule[0.1em]

$128^2\times32$ & FusedIB & - & $3\times3$ & 128 & 48 & 2 \\
\midrule[0.1em]

$64^2\times48$ & ExtraDW & $3\times3$ & $5\times5$ & 192 & 80 & 2 \\
$32^2\times80$ & ExtraDW & $3\times3$ & $3\times3$ & 160 & 80 & 1 \\
\midrule[0.1em]
$32^2\times80$ & ExtraDW & $3\times3$ & $5\times5$ & 480 & 160 & 2 \\
$16^2\times160$ & ExtraDW & $3\times3$ & $3\times3$ & 640 & 160 & 1 \\
$16^2\times160$ & ExtraDW & $3\times3$ & $3\times3$ & 640 & 160 & 1 \\
$16^2\times160$ & ExtraDW & $3\times3$ & $5\times5$ & 640 & 160 & 1 \\
$16^2\times160$ & Mobile-MQA & - & - & - & 160 & 1 \\
$16^2\times160$ & ExtraDW & $3\times3$ & $3\times3$ & 640 & 160 & 1 \\
$16^2\times160$ & Mobile-MQA & - & - & - & 160 & 1 \\
$16^2\times160$ & ConvNext & $3\times3$ & - & 640 & 160 & 1 \\
$16^2\times160$ & Mobile-MQA & - & - & - & 160 & 1 \\
$16^2\times160$ & FFN & - & - & 640 & 160 & 1 \\
$16^2\times160$ & Mobile-MQA & - & - & - & 160 & 1 \\
$16^2\times160$ & ConvNext & $3\times3$ & - & 640 & 160 & 1 \\
\midrule[0.1em]
$16^2\times160$ & ExtraDW & $5\times5$ & $5\times5$ & 960 & 256 & 2 \\
$8^2\times256$ & ExtraDW & $5\times5$ & $5\times5$ & 1024 & 256 & 1 \\
$8^2\times256$ & ExtraDW & $3\times3$ & $5\times5$ & 1024 & 256 & 1 \\
$8^2\times256$ & ExtraDW & $3\times3$ & $5\times5$ & 1024 & 256 & 1 \\
$8^2\times256$ & FFN & - & - & 1024 & 256 & 1 \\
$8^2\times256$ & ConvNext & $3\times3$ & - & 1024 & 256 & 1 \\
$8^2\times256$ & ExtraDW & $3\times3$ & $5\times5$ & 512 & 256 & 1 \\
$8^2\times256$ & Mobile-MQA & - & - & - & 256 & 1 \\
$8^2\times256$ & ExtraDW & $5\times5$ & $5\times5$ & 1024 & 256 & 1 \\
$8^2\times256$ & Mobile-MQA & - & - & - & 256 & 1 \\
$8^2\times256$ & FFN & - & - & 1024 & 256 & 1 \\
$8^2\times256$ & Mobile-MQA & - & - & - & 256 & 1 \\
$8^2\times256$ & FFN & - & - & 1024 & 256 & 1 \\
$8^2\times256$ & Mobile-MQA & - & - & - & 256 & 1 \\
$8^2\times256$ & ConvNext & $5\times5$ & - & 1024 & 256 & 1 \\
\midrule[0.1em]
$8^2\times256$ & Conv2D & - & $1\times1$ & - & 960 & 1 \\
$8^2\times960$ & AvgPool & - & $8\times8$ & - & 960 & 1 \\
$1^2\times960$ & Conv2D & - & $1\times1$ & - & 1280 & 1 \\
$1^2\times1280$ & Conv2D & - & $1\times1$ & - & 1000 & 1 \\
\bottomrule[0.2em]
\end{tabular}
\label{arch_2x_hybrid}
\end{table}
\begin{table}
\centering
\scriptsize
\caption{Architecture specification of MNv4-Conv-L.}
\begin{tabular}{c|c|c|c|c|c|c}
\toprule[0.2em]
Input & Block & DW $K_1$ & DW $K_2$ & Expanded Dim & Output Dim & Stride \\
\midrule[0.2em]
$384^2\times3$ & Conv2D & - & $3\times3$ & - & 24 & 2 \\
\midrule[0.1em]
$192^2\times24$ & FusedIB & - & $3\times3$ & 96 & 48 & 2 \\
\midrule[0.1em]
$96^2\times48$ & ExtraDW & $3\times3$ & $5\times5$ & 192 & 96 & 2 \\
$48^2\times96$ & ExtraDW & $3\times3$ & $3\times3$ & 384 & 96 & 1 \\
\midrule[0.1em]
$48^2\times96$ & ExtraDW & $3\times3$ & $5\times5$ & 384 & 192 & 2 \\
$24^2\times192$ & ExtraDW & $3\times3$ & $3\times3$ & 768 & 192 & 1 \\
$24^2\times192$ & ExtraDW & $3\times3$ & $3\times3$ & 768 & 192 & 1 \\
$24^2\times192$ & ExtraDW & $3\times3$ & $3\times3$ & 768 & 192 & 1 \\
$24^2\times192$ & ExtraDW & $3\times3$ & $5\times5$ & 768 & 192 & 1 \\
$24^2\times192$ & ExtraDW & $5\times5$ & $3\times3$ & 768 & 192 & 1 \\
$24^2\times192$ & ExtraDW & $5\times5$ & $3\times3$ & 768 & 192 & 1 \\
$24^2\times192$ & ExtraDW & $5\times5$ & $3\times3$ & 768 & 192 & 1 \\
$24^2\times192$ & ExtraDW & $5\times5$ & $3\times3$ & 768 & 192 & 1 \\
$24^2\times192$ & ExtraDW & $5\times5$ & $3\times3$ & 768 & 192 & 1 \\
$24^2\times192$ & ConvNext & $3\times3$ & - & 768 & 192 & 1 \\
\midrule[0.1em]
$24^2\times192$ & ExtraDW & $5\times5$ & $5\times5$ & 768 & 512 & 2 \\
$12^2\times512$ & ExtraDW & $5\times5$ & $5\times5$ & 2048 & 512 & 1 \\
$12^2\times512$ & ExtraDW & $5\times5$ & $5\times5$ & 2048 & 512 & 1 \\
$12^2\times512$ & ExtraDW & $5\times5$ & $5\times5$ & 2048 & 512 & 1 \\
$12^2\times512$ & ConvNext & $5\times5$ & - & 2048 & 512 & 1 \\
$12^2\times512$ & ExtraDW & $5\times5$ & $3\times3$ & 2048 & 512 & 1 \\
$12^2\times512$ & ConvNext & $5\times5$ & - & 2048 & 512 & 1 \\
$12^2\times512$ & ConvNext & $5\times5$ & - & 2048 & 512 & 1 \\
$12^2\times512$ & ExtraDW & $5\times5$ & $3\times3$ & 2048 & 512 & 1 \\
$12^2\times512$ & ExtraDW & $5\times5$ & $5\times5$ & 2048 & 512 & 1 \\
$12^2\times512$ & ConvNext & $5\times5$ & - & 2048 & 512 & 1 \\
$12^2\times512$ & ConvNext & $5\times5$ & - & 2048 & 512 & 1 \\
$12^2\times512$ & ConvNext & $5\times5$ & - & 2048 & 512 & 1 \\
\midrule[0.1em]
$12^2\times512$ & Conv2D & - & $1\times1$ & - & 960 & 1 \\
$12^2\times960$ & AvgPool & - & $12\times12$ & - & 960 & 1 \\
$1^2\times960$ & Conv2D & - & $1\times1$ & - & 1280 & 1 \\
$1^2\times1280$ & Conv2D & - & $1\times1$ & - & 1000 & 1 \\
\bottomrule[0.2em]
\end{tabular}
\label{arch_8x}
\end{table}
\begin{table}
\centering
\scriptsize
\caption{Architecture specification of MNv4-Hybrid-L.}
\begin{tabular}{c|c|c|c|c|c|c}
\toprule[0.2em]
Input & Block & DW $K_1$ & DW $K_2$ & Expanded Dim & Output Dim & Stride \\
\midrule[0.2em]
$384^2\times3$ & Conv2D & - & $3\times3$ & - & 24 & 2 \\
\midrule[0.1em]
$192^2\times24$ & FusedIB & - & $3\times3$ & 96 & 48 & 2 \\
\midrule[0.1em]
$96^2\times48$ & ExtraDW & $3\times3$ & $5\times5$ & 192 & 96 & 2 \\
$48^2\times96$ & ExtraDW & $3\times3$ & $3\times3$ & 384 & 96 & 1 \\
\midrule[0.1em]
$48^2\times96$ & ExtraDW & $3\times3$ & $5\times5$ & 384 & 192 & 2 \\
$24^2\times192$ & ExtraDW & $3\times3$ & $3\times3$ & 768 & 192 & 1 \\
$24^2\times192$ & ExtraDW & $3\times3$ & $3\times3$ & 768 & 192 & 1 \\
$24^2\times192$ & ExtraDW & $3\times3$ & $3\times3$ & 768 & 192 & 1 \\
$24^2\times192$ & ExtraDW & $3\times3$ & $5\times5$ & 768 & 192 & 1 \\
$24^2\times192$ & ExtraDW & $5\times5$ & $3\times3$ & 768 & 192 & 1 \\
$24^2\times192$ & ExtraDW & $5\times5$ & $3\times3$ & 768 & 192 & 1 \\
$24^2\times192$ & Mobile-MQA & - & - & - & 192 & 1 \\
$24^2\times192$ & ExtraDW & $5\times5$ & $3\times3$ & 768 & 192 & 1 \\
$24^2\times192$ & Mobile-MQA & - & - & - & 192 & 1 \\
$24^2\times192$ & ExtraDW & $5\times5$ & $3\times3$ & 768 & 192 & 1 \\
$24^2\times192$ & Mobile-MQA & - & - & - & 192 & 1 \\
$24^2\times192$ & ExtraDW & $5\times5$ & $3\times3$ & 768 & 192 & 1 \\
$24^2\times192$ & Mobile-MQA & - & - & - & 192 & 1 \\
$24^2\times192$ & ConvNext & $3\times3$ & - & 768 & 192 & 1 \\
\midrule[0.1em]
$24^2\times192$ & ExtraDW & $5\times5$ & $5\times5$ & 768 & 512 & 2 \\
$12^2\times512$ & ExtraDW & $5\times5$ & $5\times5$ & 2048 & 512 & 1 \\
$12^2\times512$ & ExtraDW & $5\times5$ & $5\times5$ & 2048 & 512 & 1 \\
$12^2\times512$ & ExtraDW & $5\times5$ & $5\times5$ & 2048 & 512 & 1 \\
$12^2\times512$ & ConvNext & $5\times5$ & - & 2048 & 512 & 1 \\
$12^2\times512$ & ExtraDW & $5\times5$ & $3\times3$ & 2048 & 512 & 1 \\
$12^2\times512$ & ConvNext & $5\times5$ & - & 2048 & 512 & 1 \\
$12^2\times512$ & ConvNext & $5\times5$ & - & 2048 & 512 & 1 \\
$12^2\times512$ & ExtraDW & $5\times5$ & $3\times3$ & 2048 & 512 & 1 \\
$12^2\times512$ & ExtraDW & $5\times5$ & $5\times5$ & 2048 & 512 & 1 \\
$12^2\times512$ & Mobile-MQA & - & - & - & 512 & 1 \\
$12^2\times512$ & ConvNext & $5\times5$ & - & 2048 & 512 & 1 \\
$12^2\times512$ & Mobile-MQA & - & - & - & 512 & 1 \\
$12^2\times512$ & ConvNext & $5\times5$ & - & 2048 & 512 & 1 \\
$12^2\times512$ & Mobile-MQA & - & - & - & 512 & 1 \\
$12^2\times512$ & ConvNext & $5\times5$ & - & 2048 & 512 & 1 \\
$12^2\times512$ & Mobile-MQA & - & - & - & 512 & 1 \\
$12^2\times512$ & ConvNext & $5\times5$ & - & 2048 & 512 & 1 \\
\midrule[0.1em]
$12^2\times512$ & Conv2D & - & $1\times1$ & - & 960 & 1 \\
$12^2\times960$ & AvgPool & - & $12\times12$ & - & 960 & 1 \\
$1^2\times960$ & Conv2D & - & $1\times1$ & - & 1280 & 1 \\
$1^2\times1280$ & Conv2D & - & $1\times1$ & - & 1000 & 1 \\
\bottomrule[0.2em]
\end{tabular}
\label{arch_8x_max}
\end{table}
\clearpage

\section{Larger Pareto curve}
\label{appendix:pareto_curve}
\begin{figure}[!htbp]
    \centering 
    \includegraphics[angle=270,origin=c,width=0.85\columnwidth]{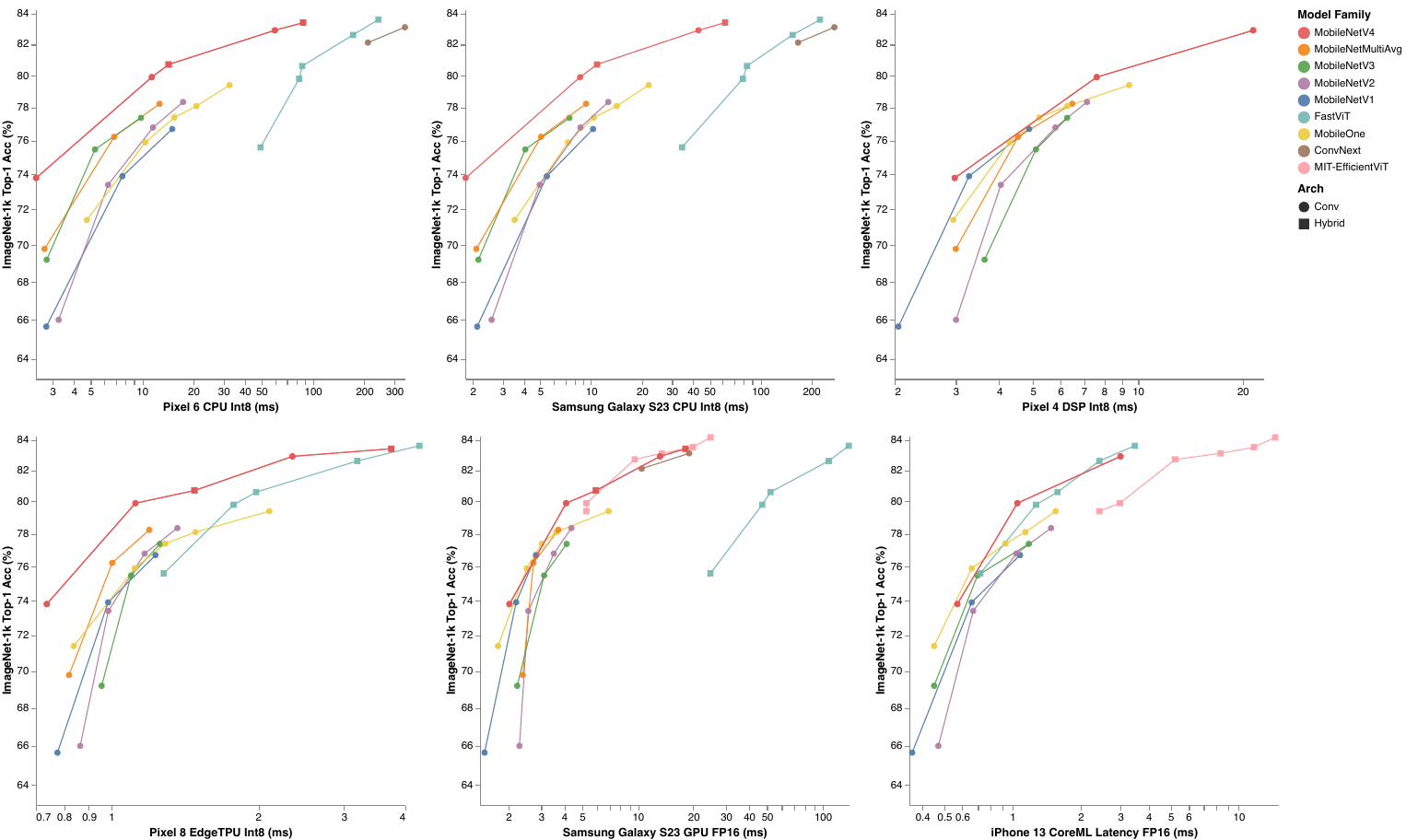}
\caption{\small{\textbf{MNv4 Models are Universally Mostly Pareto Optimal}:
This is the same chart as \cref{fig:multi_hardware_pareto}, but expanded to be easier to read.}
}
\label{fig:large_multi_hardware_pareto}
\end{figure}
\clearpage

\section{Additional Roofline Analysis}
\label{appendix:universality}
\begin{figure}
    \centering 
    \includegraphics[width=\columnwidth]{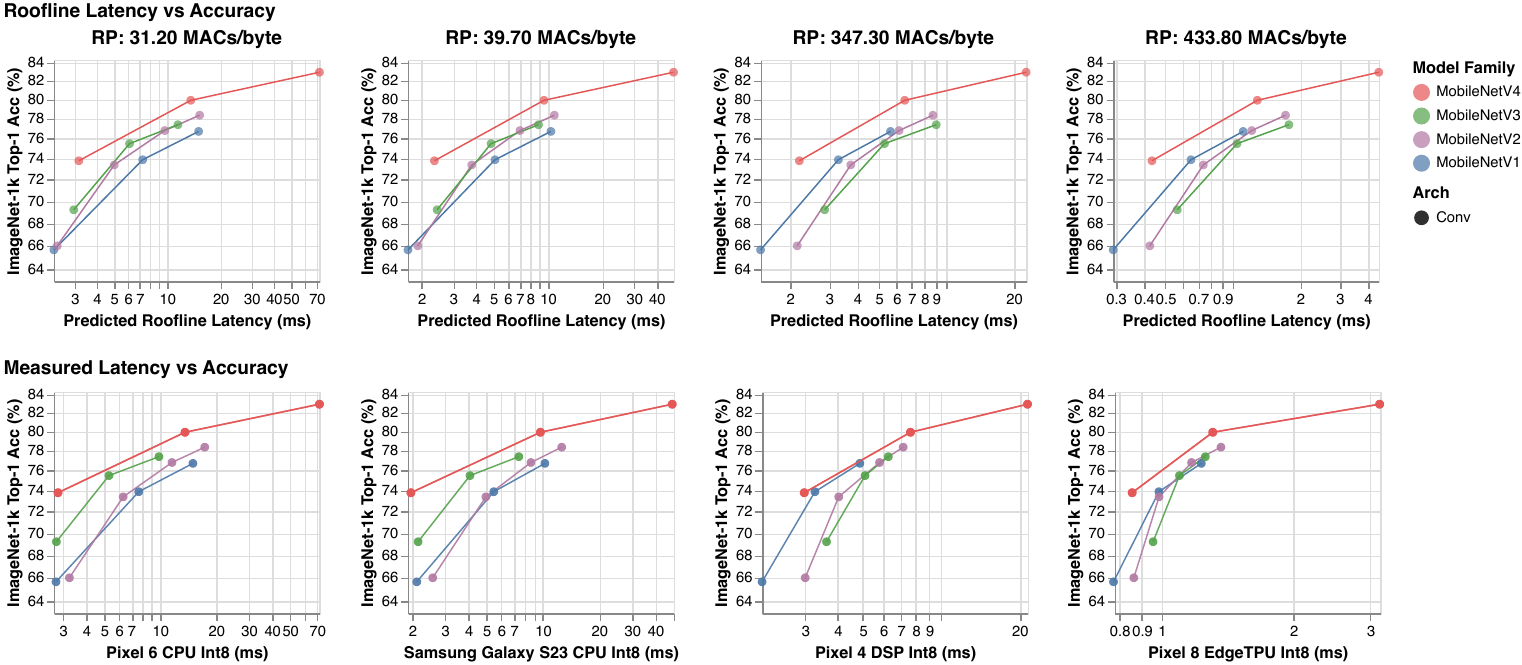}
\caption{\small{\textbf{Correlation Between Roofline Models and Real Hardware}:
These are the models considered to produce \cref{tab:roofline-correlation}}.
The roofline models successfully capture the relative ordering of each model family on each hardware target with respect to the Pareto frontier, but each target contains additional nuance that is not captured by the roofline models.}
\label{fig:roofline-correlation}
\end{figure}

\begin{figure}
    \centering 
    \includegraphics[width=\columnwidth]{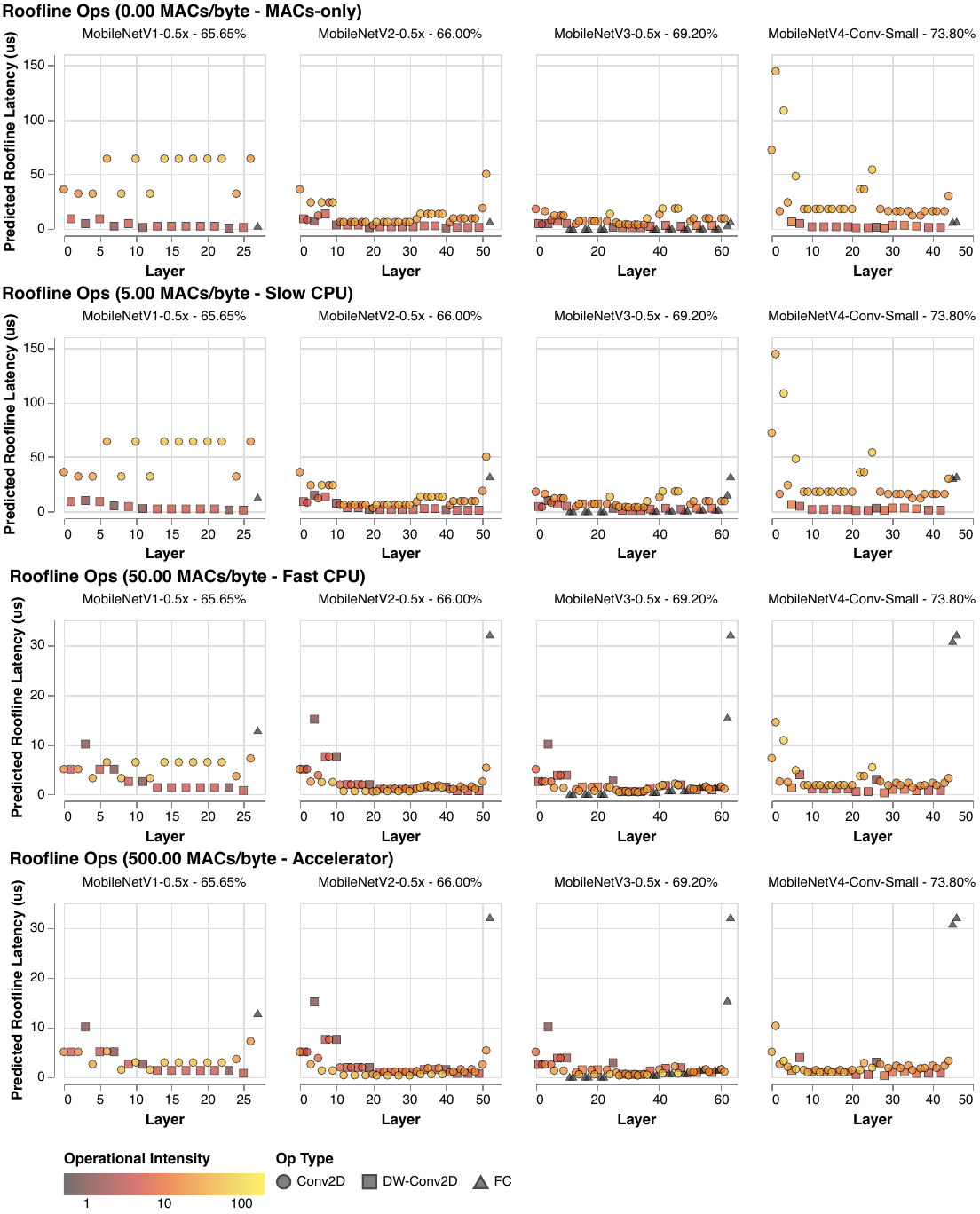}
\caption{\small \textbf{Roofline RP Sweep Analysis - Small Models}}
\label{fig:roofline-rp-sweep-small}
\end{figure}

\begin{figure}
    \centering 
    \includegraphics[width=\columnwidth]{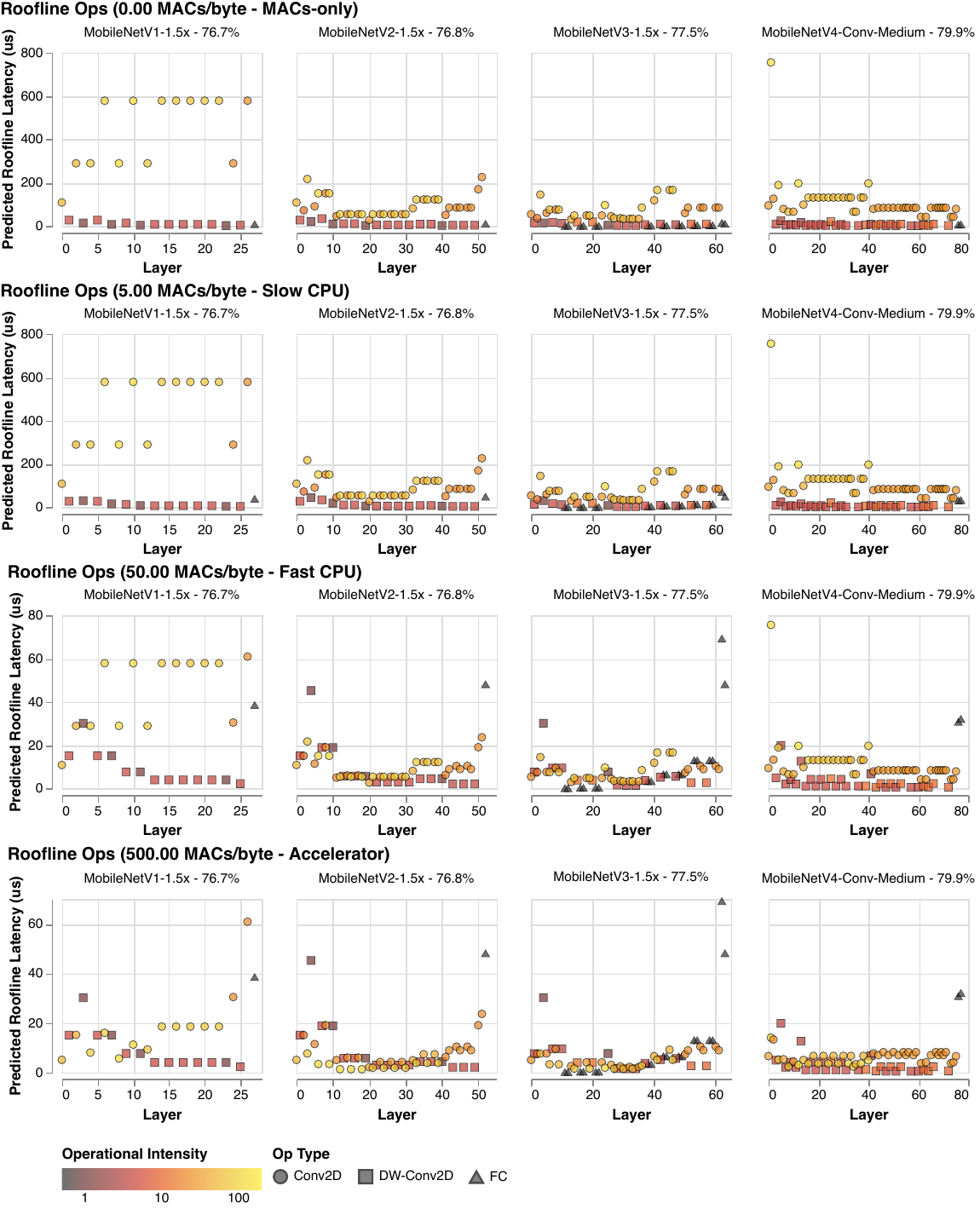}
\caption{\small \textbf{Roofline RP Sweep Analysis - Medium Models}}
\label{fig:roofline-rp-sweep-medium}
\end{figure}

\begin{figure}
    \centering 
    \makebox[\textwidth]{\includegraphics[height=.6\paperheight]{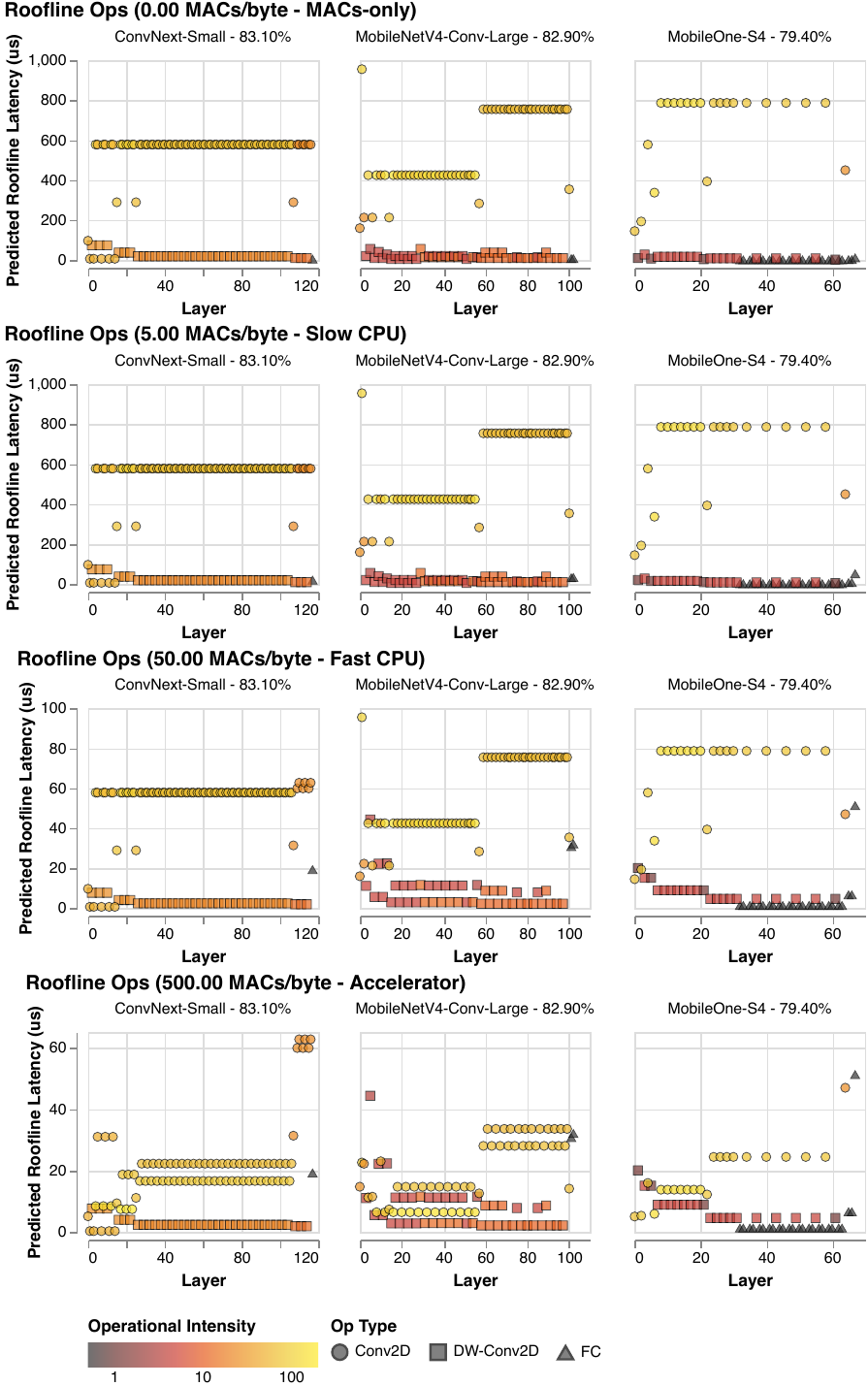}}
\caption{\small \textbf{Roofline RP Sweep Analysis - Large Models}: No previous convolutional MobileNets are as big as MobileNetV4-Conv-Large, so this compares to ConvNext-Small (\textit{left}) and MobileOne-S4 (\textit{right}) for contrast. ConvNext-Small is included because it has a similar latency to MobileNetV4-Conv-Large on S23 GPU. MobileOne-S4 is included because it has a similar latency to MobileNetV4-Conv-Large on Pixel 8 EdgeTPU.}
\label{fig:roofline-rp-sweep-large}
\end{figure}

This extends the analysis from \cref{fig:op-cost-vs-ridge-point} to include MobileNetV4-Conv-Small\\(\cref{fig:roofline-rp-sweep-small}), MobileNetV4-Conv-Medium (\cref{fig:roofline-rp-sweep-medium}), and MobileNetV4-Conv-Large (\cref{fig:roofline-rp-sweep-large}).
These figures also break out each order of magnitude in the sweep from a 0.0 MACs/byte ridge point (MACs-only, infinite memory bandwidth) to a 500.0 MACs/byte ridge point (Accelerator-like, bottlenecked on memory bandwidth).
Also included is a correlation analysis between measured latencies, empirically-fit roofline models, and counting MACs (\cref{tab:roofline-correlation} and \cref{fig:roofline-correlation}).

\begin{table}
\centering
\scriptsize
\caption{\small{\textbf{Correlation Between Roofline Models and Real Hardware}:
The Ridge Points (RPs) for these roofline models were empirically fit to the networks' measured performance.
$r_s$-Roofline is Spearman's rank correlation coefficient between the target's measured latencies and roofline predictions.
$r_s$-MAC is rank correlation between the target's measured latencies and the networks' MAC counts.
Counting MACs has high rank correlation for low-RP targets, but much lower rank correlation for high-RP targets. 
$r_s$-Roofline is high for all targets under consideration.
This shows the accuracy improvement from considering memory bandwidth in addition to MACs when estimating latency.
All Ridge Point (RP) values fit in the 0-500 MACs/B range considered in our design analysis.
These results are visualized in \cref{fig:roofline-correlation}.
}}
\begin{tabular}{l|r|c|c}
\toprule[0.2em]
                 & Ridge Point &                &          \\
Execution Target &    (MACs/B) & $r_s$-Roofline & $r_s$-MAC\\
\midrule[0.2em]
Pixel 6 CPU (Int8)            &  31.2 & 0.973 & 0.962 \\
Samsung Galaxy S23 CPU (Int8) &  39.7 & 0.962 & 0.940 \\
Pixel 4 DSP (Int8)            & 347.3 & 0.962 & 0.758 \\
Pixel 8 EdgeTPU (Int8)        & 433.8 & 0.973 & 0.857 \\
\bottomrule[0.2em]
\end{tabular}
\label{tab:roofline-correlation}
\end{table}

\section{Einsum Optimization}
\lstnewenvironment{pythoncode}{
\lstset{
    language=Python,
    basicstyle=\small\ttfamily,
    commentstyle=\color{green!50!black},
    keywordstyle=\color{blue},
    stringstyle=\color{red},
    showstringspaces=false,
    frame=single
}}{}

\label{appendix:einsum}
\begin{figure}
\begin{pythoncode}
def MQA(X, M, mask, P_q, P_k, P_v, P_o):
  Q = tf.einsum("bnd,hdk->bhnk", X, P_q)
  K = tf.einsum("bmd,dk->bmk", M, P_k)
  V = tf.einsum("bmd,dv->bmv", M, P_v)
  logits = tf.einsum("bhnk,bmk->bhnm", Q, K)
  weights = tf.softmax(logits + mask)
  O = tf.einsum("bhnm,bmv->bhnv", weights, V)
  Y = tf.einsum("bhnv,hdv->bnd", O, P_o)
\end{pythoncode}
\caption{Pseudo code of original MQA.}
\label{fig:mqa_snipet}
\end{figure}

\begin{figure}
\begin{pythoncode}
def Mobile_MQA(
    X, M, mask, P_q, P_k, P_v, P_o):
  Q = tf.einsum("bnd,dhk->bnhk", X, P_q)
  K = tf.einsum("bmd,dk->bmk", M, P_k)
  V = tf.einsum("bmd,dv->bmv", M, P_v)
  logits = tf.einsum("bnhk,bmk->bnhm", Q, K)
  weights = tf.softmax(logits + mask)
  O = tf.einsum("bnhm,bmv->bnhv", weights, V)
  Y = tf.einsum("bnhv,dhv->bnd", O, P_o)
\end{pythoncode}
\caption{Pseudo code of optimized Mobile MQA.}
\label{fig:optimized_mqa_snipet}
\end{figure}


MQA, while faster than MHSA, is still 12x slower than UIB for the same MACs. 
In the following, we investigate MQA's implementation to uncovers performance issues, and introduce Mobile MQA to improve efficiency.

Einstein summation (Einsum), extensively used in MQA and MHSA implementations across TensorFlow Keras, PyTorch, and JAX, can obscure underlying computational inefficiencies. When executed on-device, Einsum operations are decomposed into sequences of tensor transposes, reshapes, and batched matrix multiplications, aiming to minimize MACs. 
However, transposes, despite not involving MACs, are highly resource-intensive due to requiring complete tensor reads and writes in memory, significantly impacting performance.
Key inefficiencies in Einsum execution include:

\textbf{Contracted and non-contracted indices are not contiguous in the input: }
Mobile inference generally does not use a batch dimension. Accordingly an Einsum with two inputs can be translated to a matrix multiplication if both the contracted and non-contracted indices from each input are adjacent to each other. This is because adjacent indices can be very cheaply reshaped to a single index for the purpose of running the matrix multiplication and then very cheaply reshaped back. If the indices are not cleanly split into a set of contracting indices and non-contracting indices, then transposition operations are needed to bring them into this form for matrix multiplication.
In the example below, for the slow implementation, two transposes must be introduced: one to transpose $O$, and one to transpose $P_o$.
\newpage
\begin{pythoncode}
# Slow implementation: 2 tranposes needed  
Y = tf.einsum("bhnv,hdv->bnd", O, P_o)
# Fast implementation: no tranpose needed  
Y = tf.einsum("bnhv,hvd->bnd", O, P_o)
\end{pythoncode}

\textbf{Interleaving non-contracting indices or reordering indices in the output: }
In the slow implementation of the following example, the non-contracted indices $h$ and $k$ from $P_q$ are interleaved with index $n$ from $X$. As a result, two transposes will be introduced, one before the matrix multiplication to transpose $P_q$, and one afterwards to swap the indices in the output tensor.   
\begin{pythoncode}
# Slow implementation: 2 tranposes needed  
Q = tf.einsum("bnd,hdk->bhnk", X, P_q)
# Fast implementation: no tranpose needed  
Q = tf.einsum("bnd,dhk->bnhk", X, P_q)
\end{pythoncode}


In Figure~\ref{fig:optimized_mqa_snipet}, we provide the pseudo-code details of our optimized Mobile MQA and compare them with the original MQA in Figure~\ref{fig:mqa_snipet}.
Mobile MQA achieves significant efficiency improvements by addressing the two main inefficiency issues identified earlier. 
It requires only a few line of code changes to reorder the indices in the EinSum computations and weight tensor definitions in MQA to eliminate unnecessary transposes.
Table~\ref{tab:attention_ablation_w_einsum_optimization} demonstrates that we reduced the latency of MQA by threefold on Pixel 7 and by 13.8\% on Pixel 8.
Additionally, we observed an almost threefold training speedup on Google Tensor v4 using the tf.keras training framework.

\begin{table}[tb]
    \centering
    \scriptsize
\caption{\textbf{Impact of Einsum Optimization:} Percentage improvement in Mobile MQA block latency compared to vanilla MQA (without Einsum) in an MNv4-Conv-L model with attention blocks added to the last stage.}
    \begin{tabular}{ c c c c c c c}
        \toprule
        model & Top-1  & MACs & Params & \multicolumn{2}{c}{EdgeTPU}  \\
        & Acc(\%) &  (G) & (M) & Pixel 7 & Pixel 8  \\
        \midrule
        base model & 84.88 & 6.0 & 30.9 & 4.31 ms & 2.35 ms \\
        \midrule
        +3 MQA  & 85.22 & 6.5  & 34.7  & 7.00 ms  & 2.64 ms \\
        (w/o Einsum optimization) &  &  &  &  &   \\
        \midrule
        +3 Mobile MQA  & 85.24 & 6.5  & 34.7  & 5.16 ms  & 2.60 ms \\
        (w/ Einsum optimization) &  &  &  & \textbf{(-68.4\%)} & \textbf{(-13.8\%)}  \\        
        \bottomrule
    \end{tabular}
    \label{tab:attention_ablation_w_einsum_optimization}
\end{table}




\section{Implementation details of our distillation recipe}
\label{appendix:distillation}
To create the offline distillation dataset, we first augment a candidate image and then compress it using JPEG encoding. We then decode the image and run inference using the teacher. The teacher's predicted probability (the softmax output) of the 1k classes is recorded and will be used later as soft labels to train the student. We use EfficientNet L2~\cite{xie2020self} as the teacher model. This teacher model has 480 M parameters and 290 G MACs. It achieves 87.5\% top-1 accuracy on ImageNet 1k.

During student training, we disable all data augmentation except left-right flip. We also apply dropout in the final fully connected layer as the only regularization method. We use cross-entropy between the teacher's soft labels and the student's predicted class probability vectors as our training loss. We train the student with the AdamW optimizer. We use a cosine learning rate schedule with warm-up. 

Besides the significant boost in model accuracy, there are other benefits of using distillation training. For example, we found that distillation training is 2.5x faster than training on ImageNet-1k. With a batch size of 16k and 400 training epochs, distillation training on MNv4-Conv-Large takes only 2.3 hours on 128 TPU v5e. 
Finally, distillation training provides significant relief from hyper-parameter tuning.

\end{document}